\documentclass[10pt,twocolumn,letterpaper]{article}

\usepackage[pagenumbers]{cvpr} 
\usepackage[accsupp]{axessibility}
\usepackage{graphicx}
\usepackage{amsmath}
\usepackage{amssymb}
\usepackage{booktabs}
\usepackage{amssymb}
\usepackage{multirow}
\usepackage{multicol}
\usepackage{bm}
\usepackage{colortbl}
\usepackage{pifont}
 
\makeatletter
\newcommand{\thickhline}{%
    \noalign {\ifnum 0=`}\fi \hrule height 1.2pt
    \futurelet \reserved@a \@xhline
}
\definecolor{mygray}{gray}{.9}
\newcommand\ourmethod{Slide}
\usepackage{pifont}

\usepackage[pagebackref,breaklinks,colorlinks]{hyperref}

\usepackage[capitalize]{cleveref}
\crefname{section}{Sec.}{Secs.}
\Crefname{section}{Section}{Sections}
\Crefname{table}{Table}{Tables}
\crefname{table}{Tab.}{Tabs.}

\def\cvprPaperID{6486} 
\def\confName{CVPR}
\def\confYear{2023}

\begin{document}

\title{Slide-Transformer: Hierarchical Vision Transformer with Local Self-Attention}

\author{
  Xuran Pan\thanks{Equal contribution.}\qquad
  Tianzhu Ye\footnotemark[1]\qquad
  Zhuofan Xia\qquad
  Shiji Song\qquad
  Gao Huang\thanks{Corresponding author.}\\
    \normalsize{Department of Automation, BNRist, Tsinghua University}
}
\maketitle

\begin{abstract}

Self-attention mechanism has been a key factor in the recent progress of Vision Transformer (ViT), which enables adaptive feature extraction from global contexts. However, existing self-attention methods either adopt sparse global attention or window attention to reduce the computation complexity, which may compromise the local feature learning or subject to some handcrafted designs. In contrast, local attention, which restricts the receptive field of each query to its own neighboring pixels, enjoys the benefits of both convolution and self-attention, namely local inductive bias and dynamic feature selection. Nevertheless, current local attention modules either use inefficient Im2Col function or rely on specific CUDA kernels that are hard to generalize to devices without CUDA support. In this paper, we propose a novel local attention module, {\rm \textbf{Slide Attention}}, which leverages common convolution operations to achieve high efficiency, flexibility and generalizability. Specifically, we first re-interpret the {\rm column-based} Im2Col function from a new {\rm row-based} perspective and use Depthwise Convolution as an efficient substitution. On this basis, we propose a deformed shifting module based on the re-parameterization technique, which further relaxes the fixed key/value positions to deformed features in the local region. In this way, our module realizes the local attention paradigm in both efficient and flexible manner. Extensive experiments show that our slide attention module is applicable to a variety of advanced Vision Transformer models and compatible with various hardware devices, and achieves consistently improved performances on comprehensive benchmarks. Code is available at \url{https://github.com/LeapLabTHU/Slide-Transformer}.

\end{abstract}

\section{Introduction}
\begin{figure}
    \centering
    \includegraphics[width=0.88\linewidth]{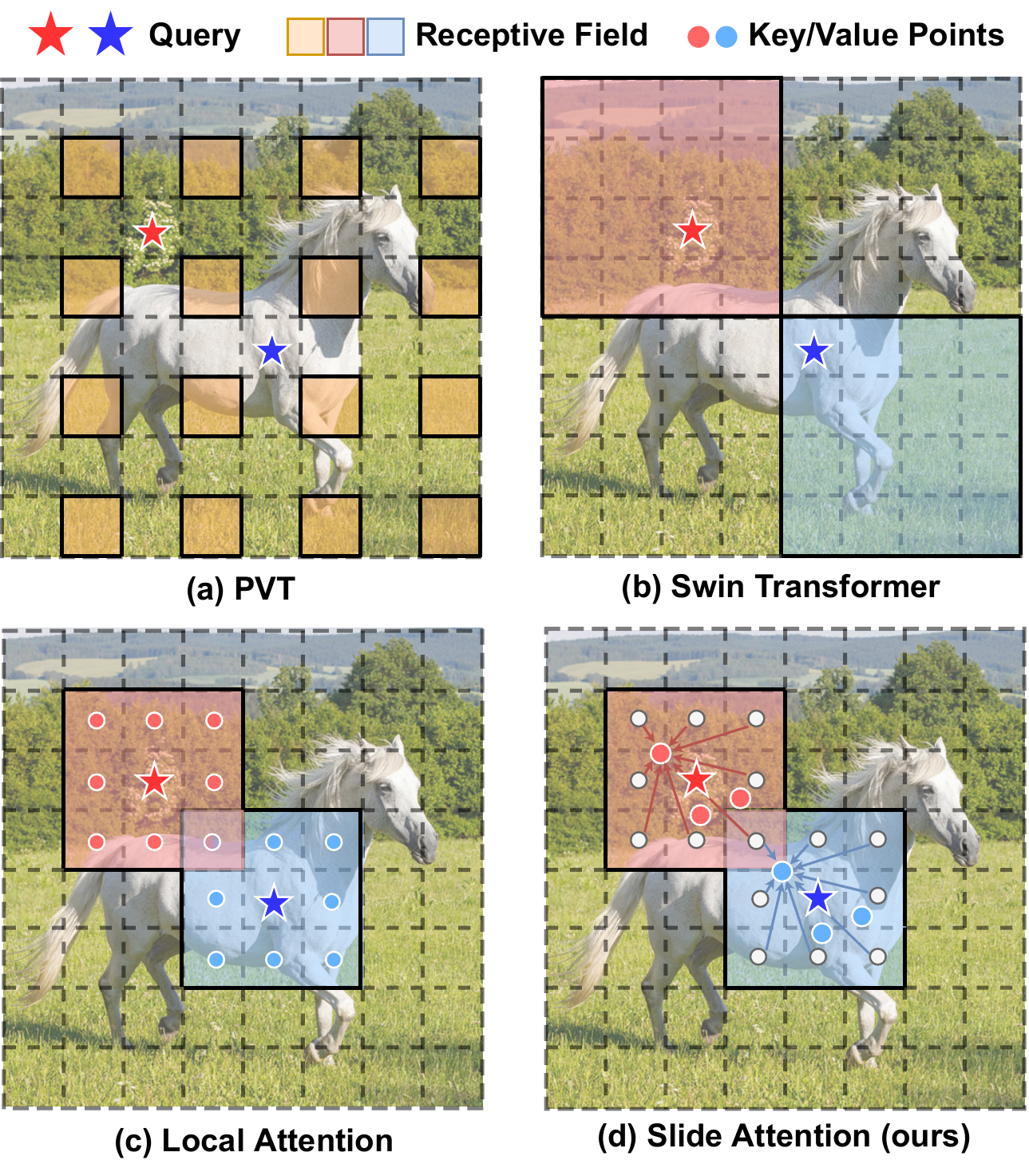}
    \vskip -0.1in
    \caption{\textbf{Comparison of our model and other attention patterns.} Comparing to the global attention in PVT and window attention in Swin-Transformer, we propose a novel Slide Attention module that not only imposes local inductive bias like local attention, but also has high efficiency and flexibility. }
    \label{fig0}
    \vskip -0.1in
\end{figure}


Transformer was originally proposed for natural language processing~\cite{attention,bert} and has gained increasing research interest in recent years. With the advent of Vision Transformer~\cite{vit}, researchers begin to realize its great potential on processing vision data, and further extend Transformer models to a variety of vision tasks including image classification~\cite{pvt,swin,acmix}, semantic segmentation~\cite{segformer,segmenter}, object detection~\cite{deformabledetr,detr,vitdet,pointformer}, and multi-modal tasks~\cite{clip,knowclip}. 


Nevertheless, adapting Transformer to vision is a non-trivial task. The computation complexity of self-attention with global receptive field grows quadratically with the sequence length, which leads to excessive computation costs and makes it impractical for vision models that require high-resolution inputs and large memory consumption.


To overcome this challenge, existing works have proposed to limit the global receptive field to smaller regions. For example, PVT~\cite{pvt} and DAT~\cite{dat} use sparse global attention to select sparse key and value positions from the feature map and share them across all queries. Another line of research including Swin Transformer~\cite{swin} and CSwin Transformer~\cite{cswin} follow the window attention paradigm. The input is divided into specially designed windows, where features are extracted and aggregated within. Despite being efficient, these carefully designed attention patterns still suffer from several limitations. On one hand, sparse global attention tends to be inferior in capturing local features, and is susceptible to key and value positions where informative features in other regions may be discarded. On the other hand, window attentions may hinder cross-window communication, and involve extra designs like window shifts that set restrictions on the model structure.

Instead of shrinking the global receptive field, a natural and effective alternative is adopting local attention by constraining receptive field of each query in its own neighboring pixels, where similar pattern has been widely used in traditional convolution design~\cite{resnet,replknet}. Compared with the aforementioned attention patterns, local attention has the advantages of convolution with translation-equivariance and local inductive bias, while also enjoying the flexibility and data-dependency of the self-attention mechanism. Several works have already investigated applying local attention to modern convolution or Transformer models. However, they either use the inefficient Im2Col function~\cite{sasa} which results in huge increase in inference time, or rely on carefully written CUDA kernels~\cite{san,nat} which restrict the applicability on devices without CUDA support. Therefore, developing a local attention module with both high efficiency and high generalizability remains challenging.


In this paper, we present a novel local attention module, dubbed \textit{Slide Attention}, that can be efficiently integrated with various Vision Transformer models and hardware devices. We target the inefficient Im2Col function that was adopted in the previous local attention module and view the process from a new perspective. Specifically, the original Im2Col generates the key and value matrix from a \textbf{column-based view}, where each column represents a local region centered at a certain position of the input. Alternatively, we re-formulate the key and value matrix from a \textbf{row-based view} and show that each row corresponds to the input feature shifted in different directions. This new insight gives us the chance to take a further step, that allows us to replace the shifting operation with carefully designed Depthwise Convolutions. In this way, the Im2Col function is replaced with standard convolution operations, which can be realized in a more efficient manner and easily implemented on different hardware devices. To further enhance flexibility, we introduce a novel deformed shifting module that relaxes fixed key and value positions (Fig.\ref{fig0}(c)) to deformed features within the local region (Fig.\ref{fig0}(d)). By using a re-parameterization technique, we effectively increase the model capacity while preserving inference efficiency.

We empirically validate our module on image classification, semantic segmentation, and object detection tasks under five advanced Vision Transformer models, and show consistent improvements over all baselines. When adopted on devices without CUDA support like Metal Performance Shader (MPS) or iPhone 12, our method also proves to be efficient. For instance, our Slide Attention based on Swin-small outperforms the vanilla Swin-base model while achieving 1.7x inference speedup on iPhone 12.

\begin{figure}
    \centering
    \includegraphics[width=0.95\linewidth]{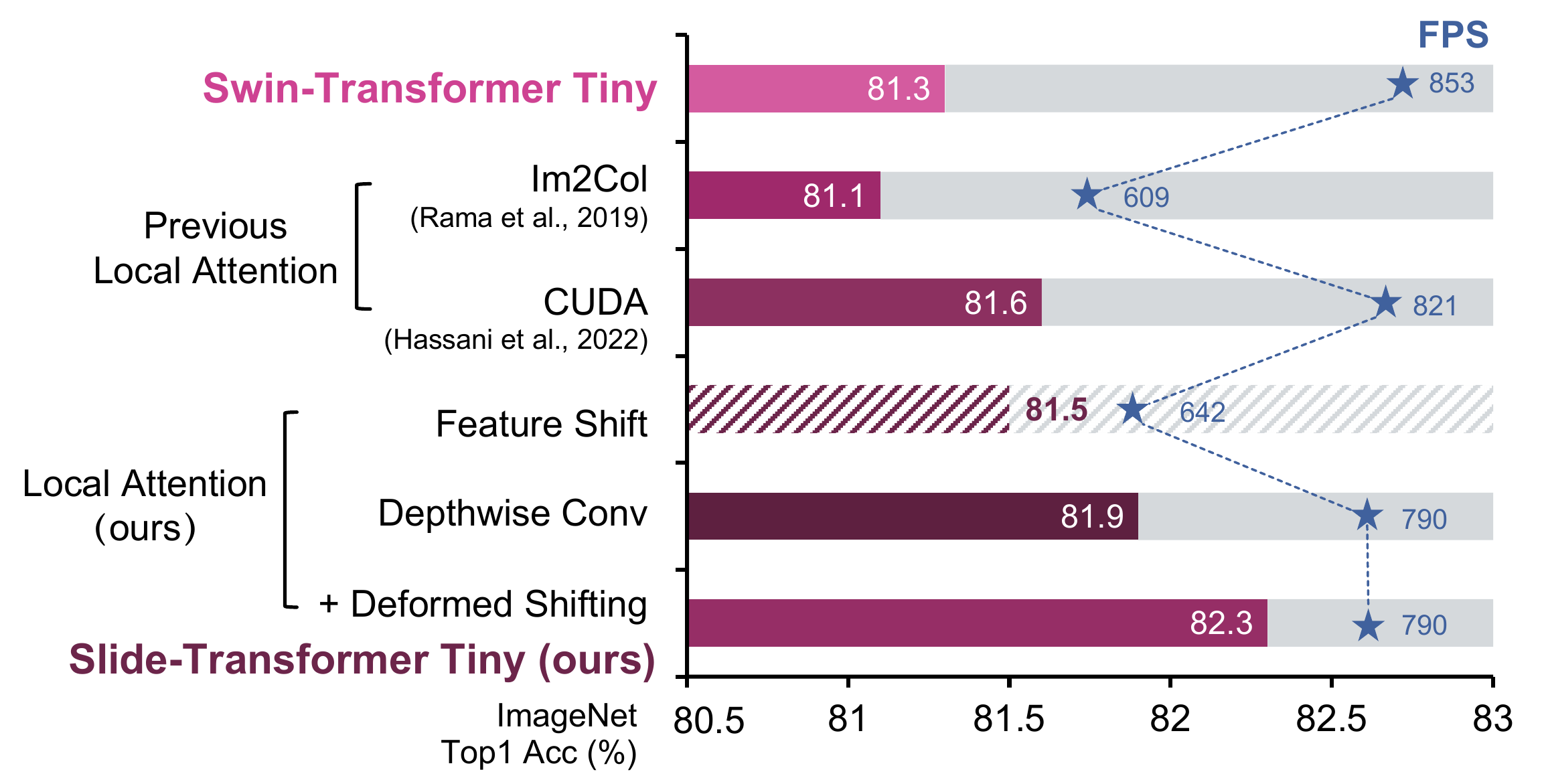}
    \vskip -0.05in
    \caption{\textbf{Performance and inference speed comparison on local attention implementations.} Results are based on Swin-Tiny~\cite{swin}. Previous works mainly use Im2Col function~\cite{sasa} or carefully designed CUDA kernels~\cite{nat}, where the former is highly inefficient and the latter only shows marginal improvements over other attention patterns, \textit{e.g.,} window attention in Swin-Transformer, and hard to generalize to other devices. Our work first re-interprets the Im2Col function as feature shift operations, then substitute shifts with more efficient depthwise convolutions. When further equipped with a deformed shifting module, our model achieves significant improvements over baselines under competitive inference time. FPS is tested on an RTX3090 GPU.}
    \label{fig1}
    \vskip -0.15in
\end{figure}

\section{Related Works}
\subsection{Vision Transformer}
Transformer and the self-attention mechanism have shown great progress in the field of Natural Language Processing~\cite{attention,bert} and successfully applied to vision tasks thanks to the pioneering work of Vision Transformer~\cite{vit}. Following its path, researchers have extended Vision Transformer models along various directions, including data efficiency~\cite{deit}, position encoding~\cite{position}, and optimization~\cite{early}. To better adapt Vision Transformers to downstream tasks, several works focused on investigating pyramid model structures, and show advanced performances over convolution-based approaches. PVT~\cite{pvt,pvtv2} considers sampling sparse locations in the feature map as key and value pairs. DAT~\cite{dat} takes a further step and shifts fixed locations toward different directions in a data-dependent way. MViT~\cite{mvit,mvitv2} considers the pooling function on the input to obtain key and value pairs, which can be seen as a lower resolution of the feature map. Other approaches adopt an alternative strategy and restrict the attention to carefully designed patterns. Swin Transformer~\cite{swin} designs non-overlapped windows and shifts windows between consecutive blocks. On this basis, CSwin Transformer~\cite{cswin} adopts a cross-shape window to further improve model capacity.

\subsection{Local Attention}
By constraining the attention receptive field of each query in its own neighboring pixels, local attention inherits the advantages from traditional convolution including local inductive bias and translation-equivariance~\cite{sasa}. 
Researchers follow this path and target improving the efficiency of local attention. HaloNet~\cite{halonet} combines window attention with local attention by first dividing the input into blocks and considering neighborhood windows instead of pixels. Another direction is to design CUDA kernels with high inference speed. SAN~\cite{san} designs a novel patchwise attention pattern and achieves better performances based on convolution architectures. NAT~\cite{nat} adopts neighborhood attention and specifically considers situations for corner pixels. Nevertheless, current local attention models either use inefficient Im2Col function and endure huge increase in inference time, or rely on carefully written CUDA kernels that restrict applicability on CUDA-free devices.

\section{Overview of Self-Attention}
In this section, we first provide an overview of the self-attention module and its various forms. Compared to the widely used sparse global attention and window attention paradigm, local attention tends to be the most natural implementation while suffering from efficiency limitations.
\subsection{Multi-Head Self-Attention}
Multi-head self-attention (MHSA) is the core component of Transformer models, which is also the most distinct part among the numerous Transformer researches. In general, an MHSA block with $M$ heads can be formulated as:
\begin{align}
    &q=xW_q,\ k=xW_k,\ v=xW_v, \label{eq:proj} \\
    &z^{(m)}=\sigma(q^{(m)} \cdot k^{(m)\top}_{[r_{q}]}/\sqrt{d})\cdot v^{(m)}_{[r_{q}]}, m\!=\!1,\ldots,M, \label{eq:attn}\\
    &z=\text{Concat}\left(z^{(1)},\ldots,z^{(M)}\right)W_o, \label{eq:MHSA}
\end{align}
where $\sigma(\cdot)$ denotes the SoftMax function, and $d$ is
the channel dimension of each head. In particular, we denote $r_q$ as the receptive field of a specific query $q$, and denote $k^{(m)}_{[r_{q}]}$ and $v^{(m)}_{[r_{q}]}$ as the corresponding key and value pairs respectively.

\subsection{Attention Patterns}
The implementation of self-attention in the field of computer vision is never a trivial task. Like a coin has two sides, the high flexibility of the self-attention mechanism leads to higher computation complexity and lower efficiency on hard-wares. Therefore, to achieve a better trade-off between performance and efficiency, previous works have investigated injecting different inductive biases into vanilla self-attention paradigm by designing different attention patterns.

\textbf{(1) Sparse Global Attention} ~\cite{pvt,dat} considers selecting a sparse set of key and value pairs instead of the dense feature map. However, this also restricts the potential of feature extraction into a limited subset of input. Also, the key and value pairs are the same for all queries. This query-agnostic selection strategy may lead to a homogenization of features throughout the whole feature map.


\textbf{(2) Window Attention} ~\cite{swin,cswin} is another option to carefully divide input into particular windows where features are extracted within. Although partially addressing the limitation of query-agnostic key and value pairs, the designed patterns may lead to unnatural circumstances where features at the edge of different windows are totally isolated despite being close in the feature map. Also, window patterns need to shift between consecutive blocks to facilitate connections across windows, involving extra designs in model structure.

\textbf{(3) Local Attention} constrains the receptive field of each query in its own neighboring pixels, sharing a similar pattern with convolution. Compared to former patterns, local attention enjoys the advantages from both convolution and self-attention: 1) Local inductive bias from a query-centric attention pattern;
2) Translation-equivariance like traditional convolution, showing robustness towards shift variances of input; 3) Involving little human design, which sets the least restrictions on the model architecture design. 

\begin{figure*}
    \centering
    \includegraphics[width=0.88\linewidth]{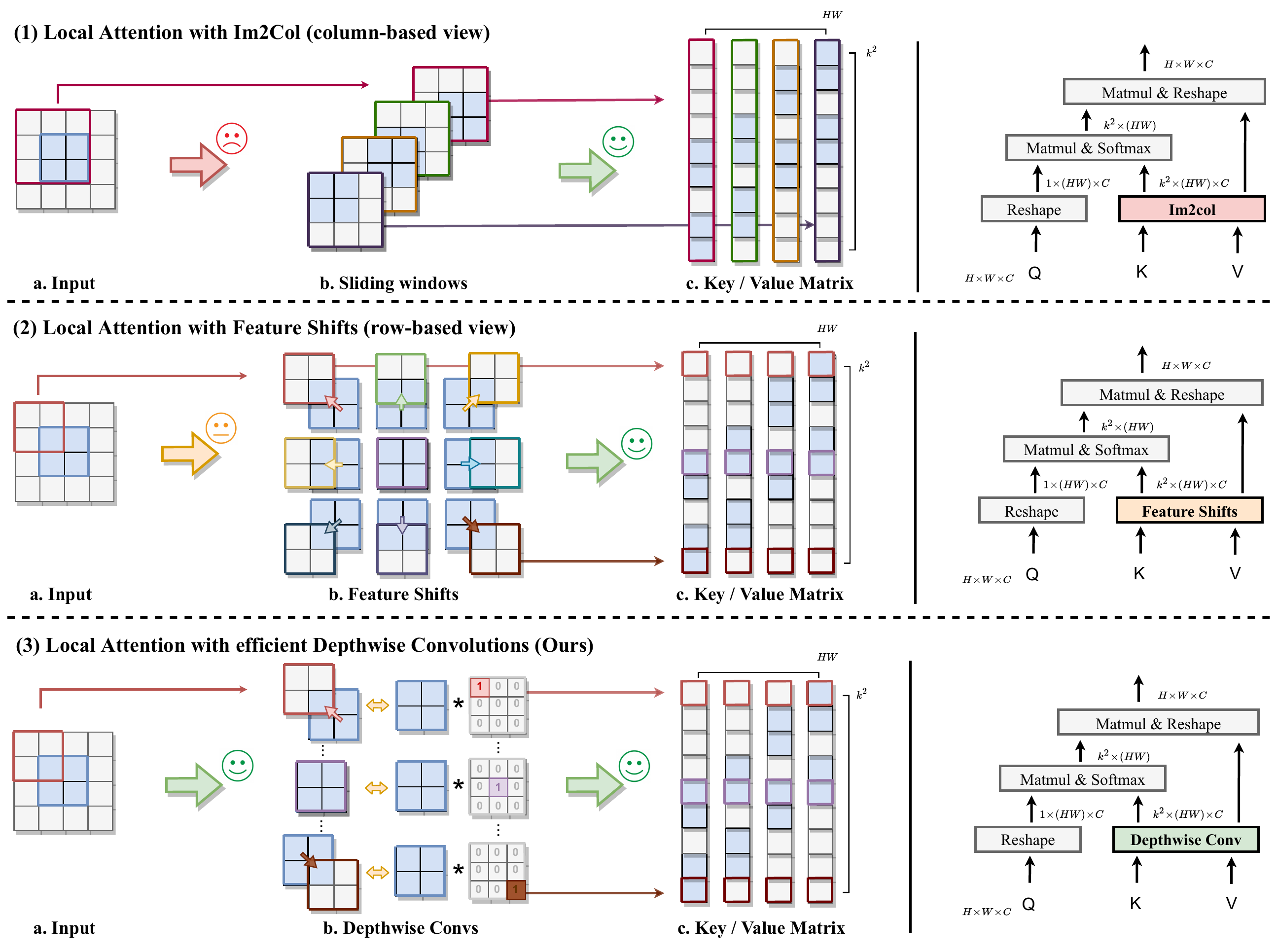}
    \vskip -0.1in
    \caption{\textbf{Different implementation on the local attention module.} We take 3x3 local attention on a 2x2 feature map (in blue) with [1,1] padding (in gray) as an example. \textbf{Sub-figure(1)}: Im2Col function is viewed in a \textit{column-based} way, where each column of the key/value matrix corresponds to the local region of a particular query (1.b). The process of sampling windows breaks data locality and leads to inefficiency \textcolor{red}{\ding{55}}. \textbf{Sub-figure(2)}: we view the key/value matrix in a \textit{row-based} way, where each row is equivalent to the input feature, only after shifting towards certain directions (2.b). Nevertheless, shifting toward different directions is also inefficient when compared with common operators \textcolor{yellow}{\ding{55}}. \textbf{Sub-figure(3)}: we take a step forward, and substitute shifting operations with carefully designed depthwise convolutions, which is not only efficient but also friendly to different hardware implementations \textcolor{green}{\ding{51}}. Best viewed in color.}
    \label{fig:fig2}
    \vskip -0.1in
\end{figure*}

\subsection{Local Attention Implementation}
\label{local}
Despite being effective, the local receptive field also poses difficulties in practical implementation. Specifically, due to the fact that the receptive region is different for each query in the feature map, special technique, \textit{i.e.,} Im2Col function needs to be adopted to sample keys and values for all the queries respectively. 
As illustrated in Fig.\ref{fig:fig2}(1), local window is centered at a particular query and represents the region of its corresponding key/value pairs. The windows are then flattened into \textbf{columns} and consist of the final key/value matrix. However, the process of sampling windows is mainly achieved by independently slicing the feature map, which practically breaks data locality and leads to huge time consumption. In the case of convolution, special tricks like Winograd~\cite{winograd} can be adopted where a portion of computations can be pre-computed before the inference stage. However, the tricks fail to generalize to local attention, since the \textit{'kernel weights'} are computed by the dot product of queries and keys in a data-dependent way.



Another line of research~\cite{san,nat} focuses on improving the efficiency of local attention by writing CUDA kernels to replace the inefficient Im2Col function. Albeit effective, this inevitably sets restrictions on the potential applicability, making it impractical on hardwares without CUDA support, especially on edge devices like smartphones.

To show a thorough comparison between the aforementioned approaches, we practically analyze the performance and runtime of these two implementations of local self-attention and compare it with the window attention in Swin-Transformer. As illustrated in Fig.\ref{fig1}, Im2Col-based local attention is less favorable in both efficiency and performance. The CUDA-based approaches can maintain comparable inference speed with vectorized operations like window attention, while only achieving marginal improvements. Considering the difficulty of adopting CUDA kernels on different hardwares, we still lack a local attention module that has both high efficiency and high generalizability.

\section{Method}
As analyzed above, local attention suffers from the efficiency problem that prevents it from practical implementation. In this section, we first show that the inefficient Im2Col function can be re-interpreted from another perspective and proved to be equivalent to a group of simple feature shifts. On this basis, we substitute feature shift operations with efficient depthwise convolutions. Equipped with a novel deformed shifting module to relax the fixed local key/value positions to deformed features, we finally propose a local attention module, dubbed \textit{Slide Attention}, with high efficiency and flexibility.
\subsection{New Perspective on Im2Col}
\label{new}
For better understanding, we first review the process of the Im2Col function. We take the operations on keys as an example in the following section, and the case for values is exactly the same. Let ${\bf K}\in \mathcal{R}^{H\!\times\!W\!\times\!C}$ denote the keys of the self-attention module and $k$ denote the local window size, the output of Im2Col can be represented as:
\begin{align}
    &O_k[u*k\!+\!v,i*H\!+\!j] = {\bf K}[i+u,j+v], \\
    {\rm for} \ i\!\in\!&[0,W\!\!-\!\!1], j\!\in\!\![0,H\!\!-\!\!1], u,v\!\in\![\!-\!\lfloor{k/2}\rfloor,\lfloor{k/2}\rfloor].
\end{align}
From the \textbf{column-based} view, as illustrated in Fig.\ref{fig:fig2}[1], the key/value matrix contains $HW$ columns where each column corresponds to a local window centered at a particular query.
Specifically, if we carefully check each column of the output, the above equations can be reformulated as:
\begin{align}
    O_k[:,&i*H\!+\!j] = {\rm Column}^{(i,j)}, \\
    &{\rm where} \ \ {\rm Column}^{(i,j)}(u,v)\!=\! {\bf K}[i+u,j+v]
\end{align}
represents a local window centered at $(i,j)$. 
This is in accordance with motivation of Im2Col function, where receptive windows of all queries are sampled and placed in order.

However, an interesting observation is we can also view the Im2Col function in a different way. From the \textbf{row-based} view, as illustrated in Fig.\ref{fig:fig2}(2), the key/value matrix contains $k^2$ rows, where each row corresponds to shifting input towards a certain direction. Specifically, we focus on each row of the output 
and reformulate the above equations as:
\vskip -0.2in
\begin{align}
    O_k[u*&k\!+\!v,:] = {\rm Row}^{(u,v)}, \\
    &{\rm where} \ \ {\rm Row}^{(u,v)}(i,j)= {\bf K}[i+u,j+v] \label{column_view}
\end{align}
is equivalent to shifting the original feature map towards a certain direction $(u,v)\!\in\![\!-\!\lfloor{k/2}\rfloor,\lfloor{k/2}\rfloor]$.

In this way, we offer a new alternative to understanding the Im2Col function by substituting the \textbf{column-based} view with a novel \textbf{row-based} view. Take $k=3$ as an example, if we first shift the original feature map towards 9 different directions (Fig.\ref{fig:fig2}(2.b)), then flatten these features into rows and finally concatenate them in column (Fig.\ref{fig:fig2}(2.c)), the obtained key/value matrix is proved equivalent to $HW$ local windows which can recover the exact same output of the original Im2Col function (Fig.\ref{fig:fig2}(1.c)).

\subsection{Shift as Depthwise Convolution}
\label{shift}
Although the re-interpretation in Sec.\ref{new} provides us a new way to understand the Im2Col function, simply shifting features towards different directions still involves inefficient slicing operations, which provide little help in promoting the efficiency of local attention. Nevertheless, unlike sampling windows of all queries in Im2Col, feature shifting can be achieved in a more efficient way.

Specifically, we resort to applying depthwise convolution with designed kernels as a replacement for the inefficient feature shifts, as shown in Fig.\ref{fig:fig2}(3). Take $u\!=\!-1$ and $v\!=\!-1$ in Eq.(\ref{column_view}) as an example, for a input $f\in\mathcal{R}^{H\!\times\!W\!\times\!C}$, shifting towards direction $(-1,-1)$ can be formulated as:
\begin{equation}
    \label{shift_ex}
    \tilde{f}[i,j,:] = f[i-1,j-1,:], \ \forall i,j.
\end{equation}
On the other hand, if we denote the depthwise convolution kernel (kernel size $k=3$) as:
\begin{equation}
\setlength{\abovedisplayskip}{1ex}
    K[:,:,c] = \left[
    \begin{array}{ccc}
    1 & 0 & 0 \\
    0 & 0 & 0 \\
    0 & 0 & 0 \\
    \end{array}
    \right],  \ \forall c,
\setlength{\belowdisplayskip}{1ex}
\end{equation}
the corresponding output can be formulated as:
\begin{align}
    f^{\rm (dwc)}[i,j,c] &=\!\!\!\!\!\!\!\! \sum_{p, q \in \{0,1,2\}}\!\!\!\!\!\!\! K[p,q,c]f[i\!\!+\!\!p\!\!-\!\!1, j\!\!+\!\!q\!\!-\!\!1, c] \\
    &= f[i\!\!-\!\!1, j\!\!-\!\!1,c] = \tilde{f}[i,j,c], \ \forall i,j,c.
\end{align}
Therefore, with carefully designed kernel weights for different shift directions, the convolution outputs are equivalent to the simple feature shifts.

In general, we can integrate the findings from Sec.\ref{new} and Sec.\ref{shift} and propose an efficient implementation of local attention. For local attention with window size $k$, we can re-implement the Im2Col function as $k^2$ carefully defined depthwise convolutions, alleviating the main overhead. Moreover, these depthwise convolutions can be further boiled down to a single group convolution, which not only avoids the inefficient slicing operation, but also can benefit from the optimized implementation of convolution operations on many hardwares~\cite{winograd,replknet}.

\begin{figure}
    \centering
    \includegraphics[width=0.85\linewidth]{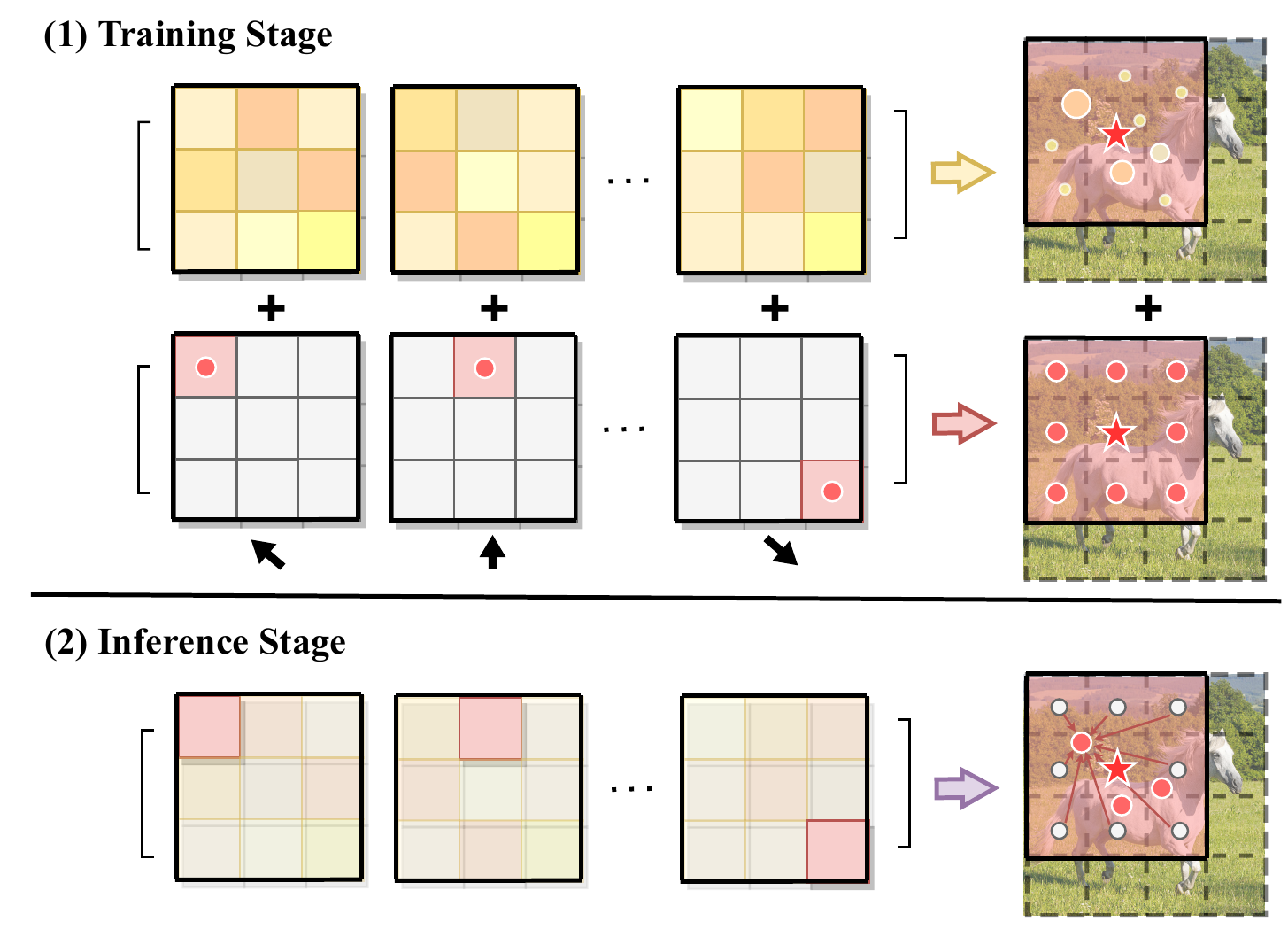}
    \vskip -0.1in
    \caption{\textbf{Deformed shifting module with re-parameterization.} (1) At the training stage, we maintain two paths, one with designed kernel weights to perform shifting towards different directions, and the other with learnable parameters to enable more flexibility. (2) At the inference stage, we merge these two convolution operations into a single path with re-parameterization, which improves the model capacity while maintaining the inference efficiency.}
    \label{fig:fig4}
    \vskip -0.15in
\end{figure}

\begin{figure*}
\begin{minipage}{0.66\linewidth}
 \centerline{\includegraphics[width=1.0\linewidth]{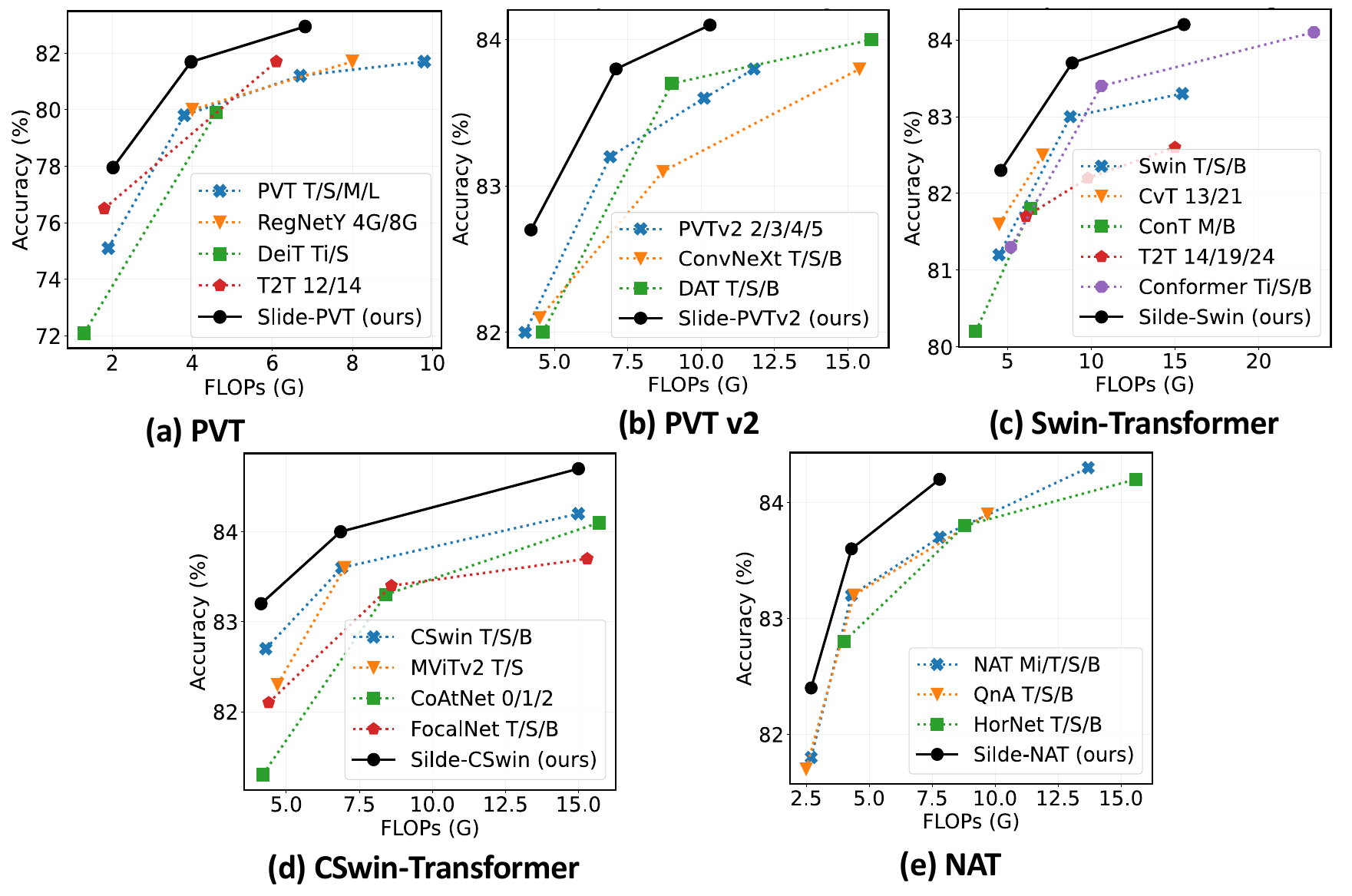}}
\end{minipage}
\hfill
\begin{minipage}{.32\linewidth}\footnotesize
\renewcommand\arraystretch{0.85}
\setlength{\tabcolsep}{1.0mm}{
\begin{tabular}{l|cc|l}
\toprule
\textbf{Method} & \textbf{Params} & \textbf{Flops} & \textbf{Top-1}\\
\midrule
PVT-T~\cite{pvt}  & 13.2M & 1.9G & 75.1\\
\rowcolor{mygray}
\textbf{Slide-PVT-T} & 12.2M & 2.0G & \textbf{78.0\,{\scriptsize (+2.9)}}\\
PVT-S & 24.5M & 3.8G & 79.8\\
\rowcolor{mygray}
\textbf{Slide-PVT-S} & 22.7M & 4.0G & \textbf{81.7\,{\scriptsize (+1.9)}}\\
\midrule
PVTv2-B1~\cite{pvtv2}  & 13.1M & 2.1G & 78.7\\
\rowcolor{mygray}
\textbf{Slide-PVTv2-B1} & 13.0M & 2.2G & \textbf{79.5\,{\scriptsize (+0.7)}}\\
PVTv2-B2  & 25.4M & 4.0G & 82.0\\
\rowcolor{mygray}
\textbf{Slide-PVTv2-B2} & 22.8M & 4.2G & \textbf{82.7\,{\scriptsize (+0.7)}}\\
\midrule
Swin-T~\cite{swin}  & 29M & 4.5G & 81.3\\
\rowcolor{mygray}
\textbf{Slide-Swin-T} & 29M & 4.6G & \textbf{82.3\,{\scriptsize (+1.0)}}\\
Swin-S & 50M & 8.7G & 83.0\\
\rowcolor{mygray}
\textbf{Slide-Swin-S} & 51M & 8.9G & \textbf{83.7\,{\scriptsize (+0.7)}}\\
Swin-B & 88M & 15.4G & 83.5\\
\rowcolor{mygray}
\textbf{Slide-Swin-B} & 89M & 15.5G & \textbf{84.2\,{\scriptsize (+0.7)}}\\
\midrule
CSwin-S~\cite{cswin} & 35M & 6.9G & 83.6\\
\rowcolor{mygray}
\textbf{Slide-CSwin-S} & 35M & 6.9G & \textbf{84.0\,{\scriptsize (+0.4)}}\\
CSwin-B & 78M & 15.0G & 84.2\\
\rowcolor{mygray}
\textbf{Slide-CSwin-B} & 78M & 15.0G & \textbf{84.7\,{\scriptsize (+0.5)}}\\
\midrule
NAT-T~\cite{cswin} & 28M & 4.3G & 83.2\\
\rowcolor{mygray}
\textbf{Slide-NAT-T} & 28M & 4.3G & \textbf{83.6\,{\scriptsize (+0.4)}}\\
NAT-S & 51M & 7.8G & 83.7\\
\rowcolor{mygray}
\textbf{Slide-NAT-S} & 51M & 7.8G & \textbf{84.3\,{\scriptsize (+0.6)}}\\
\bottomrule
\end{tabular}}
\end{minipage}
\vskip -0.1in
\caption{Comparisons of FLOPS and parameters against accuracy on ImageNet classification task. 
Models in (a)(b) adapt from PVT and PVTv2 with global attention; Models in (c)(d) adapt from Swin-Transformer and CSwin-Transformer with window attention; Models in (e) adapt from NAT with local attention. See the full comparison table in Appendix.}
\label{main}
\vskip -0.15in
\end{figure*}

\subsection{Deformed Shifting Module}
By switching the original Im2Col function to depthwise convolutions, the efficiency of the local attention is greatly improved. Nevertheless, the carefully designed kernel weights still constrain keys and values to the fixed neighboring positions, which may not be the optimal solution for capturing diverse features.


Therefore, we propose a novel \textit{deformed shifting module} to further enhance the flexibility of local attention. In specific, we take advantage of design paradigm in our shift-wise convolution, and introduce a parallel convolution path where the kernel parameters are randomly initialized and learnable in the training process. Compared to fixed kernels that shift features towards different directions, learnable kernels can be interpreted as a linear combination of all local features. This is in analogy with the deformed receptive field in Deformable Convolutional Network~\cite{dcn}, where our module practically relaxes the fixed key and value positions to deformed features in the local region.

As illustrated in Fig.\ref{fig:fig4}, the additional convolution path improves local attention module from several perspectives: 

(1) The key and value pairs in the local attention are extracted by a more flexible module, that greatly improves model capacity and can capture diverse features.

(2) The learnable convolution kernel shows a resemblance with the deformable technique in DCN. Similar to the bilinear interpolation of four neighboring pixels in DCN, our deformed shifting module can be viewed as a linear combination of features within the local window. This finally contributes to augmenting the spatial sampling locations and model geometric transformation of inputs.

(3) We use the re-parameterization technique~\cite{repvgg} to transform the two parallel paths into a single convolution. In this way, we can improve the model capacity while maintaining inference efficiency.

\subsection{Implementation}
On the basis of the aforementioned design, we propose a novel \textit{Slide Attention} module that enables a highly efficient and flexible local attention pattern and poses little restriction on model architecture design. Our block can serve as a plug-in module and is easily adopted on a variety of modern vision Transformer architectures and hardware devices. As a showcase, we empirically implement our module on five advanced models including PVT~\cite{pvt}, PVT-v2~\cite{pvtv2}, Swin Transformer~\cite{swin}, CSwin Transformer~\cite{cswin} and NAT~\cite{nat}, and conduct experiments on several environments including Nvidia GPU, Medal Performance Shader and iPhone 12.

Also, previous works~\cite{early} have demonstrated that the locality and translation-equivariant property in convolutions are beneficial at early stages of vision Transformers. Considering the similar design pattern and characteristics between our module and traditional convolution, we simply adopt the slide attention block at the early stages of vision Transformer models, and keep the rest of the block unchanged. The detailed architectures are shown in Appendix.

\begin{table*}[t]\footnotesize
\begin{center}
\setlength{\tabcolsep}{1.8mm}{
\renewcommand\arraystretch{1.05}
\begin{tabular}{l|c|c|c|ccc|ccc|ccc|ccc}
\thickhline
\multicolumn{16}{c}{\textbf{(a) Mask R-CNN Object Detection \& Instance Segmentation on COCO}} \\
Method & FLOPs & \#Param & Schedule & AP$^b$ & AP$^b_\text{50}$ & AP$^b_\text{75}$ & AP$^b_s$ & AP$^b_m$ & AP$^b_l$ & AP$^m$ & AP$^m_\text{50}$ & AP$^m_\text{75}$ & AP$^m_s$ & AP$^m_m$ & AP$^m_l$ \\
\hline
PVT-T & 240G & 33M & 1x & 36.7 & 59.2 & 39.3 & 21.6 & 39.2 & 49.0 & 35.1 & 56.7 & 37.3 & 19.5 & 37.4 & 48.5 \\
\rowcolor{mygray}
\textbf{\ourmethod-PVT-T} & 219G & 32M & 1x & 40.4 & 63.4 & 43.8 & 25.3 & 42.8 & 53.0 & 38.1 & 60.4 & 41.0 & 20.0 & 40.1 & 55.2 \\
\hline
PVT-S & 305G & 44M & 1x & 40.4 & 62.9 & 43.8 & 22.9 & 43.0 & 55.4 & 37.8 & 60.1 & 40.3 & 20.4 & 40.3 & 53.6 \\
\rowcolor{mygray}
\textbf{\ourmethod-PVT-S} & 269G & 42M & 1x & 42.8 & 65.9 & 46.7 & 26.6 & 45.5 & 57.3 & 40.1  & 63.1 & 43.1 & 20.3 & 42.4 & 59.0 \\
\hline
PVT-M & 392G & 64M & 1x & 42.0 & 64.4 & 45.6 & 24.4 & 44.9 & 57.9 & 39.0 & 61.6 & 42.1 & 21.3 & 42.0 & 55.2 \\
\rowcolor{mygray}
\textbf{\ourmethod-PVT-M} & 357G & 62M & 1x & 44.4 & 66.9 & 48.6 & 28.9 & 47.0 & 59.4 & 40.8  & 63.9 & 43.8 & 25.0 & 43.5 & 55.9 \\
\hline
PVTv2-B1 & 244G & 34M & 1x & 41.8 & 64.3 & 45.9 & 26.4 & 44.9 & 54.3 & 38.8 & 61.2 & 41.6 & 20.2 & 41.3 & 56.1 \\
\rowcolor{mygray}
\textbf{\ourmethod-PVTv2-B1} & 222G & 33M & 1x & 42.6 & 65.3 & 46.8 & 27.4 & 45.6 & 55.7 & 39.7 & 62.6 & 42.6 & 24.1 & 42.9 & 53.7 \\
\hline
PVTv2-B2 & 309G & 45M & 1x & 45.3 & 67.1 & 49.6 & 28.8 & 48.4 & 59.5 & 41.2 & 64.2 & 44.4 & 22.0 & 43.7 & 59.4 \\
\rowcolor{mygray}
\textbf{\ourmethod-PVTv2-B2} & 274G & 43M & 1x & 46.0 & 68.2 & 50.3 & 28.8 & 49.4 & 61.0 & 41.9 & 65.1 & 45.4 & 24.6 & 45.2 & 57.2 \\
\hline
Swin-T & 267G & 48M & 3x & 46.0 & 68.1 & 50.3 & 31.2 & 49.2 & 60.1 & 41.6 & 65.1 & 44.9 & 25.9 & 45.1 & 56.9 \\
\rowcolor{mygray}
\textbf{\ourmethod-Swin-T} & 268G & 48M & 3x & 46.8 & 69.0 & 51.6 & 31.7 & 50.4 & 60.1 & 42.3 & 66.0 & 45.8 & 23.5 & 45.8 & 60.8 \\
\thickhline
\multicolumn{16}{c}{\textbf{(b) Cascade Mask R-CNN Object Detection \& Instance Segmentation on COCO}} \\
Method & FLOPs & \#Param & Schedule & AP$^b$ & AP$^b_\text{50}$ & AP$^b_\text{75}$ & AP$^b_s$ & AP$^b_m$ & AP$^b_l$ & AP$^m$ & AP$^m_\text{50}$ & AP$^m_\text{75}$ & AP$^m_s$ & AP$^m_m$ & AP$^m_l$ \\
\hline
Swin-T & 745G & 86M & 3x & 50.4 & 69.2 & 54.7 & 33.8 & 54.1 & 65.2 & 43.7 & 66.6 & 47.3 & 27.3 & 47.5 & 59.0 \\
\rowcolor{mygray}
\textbf{\ourmethod-Swin-T} & 747G & 86M & 3x & 51.1 & 69.8 & 55.4 & 35.2 & 54.4 & 65.8 & 44.3 & 67.4 & 48.0 & 28.0 & 48.0 & 59.2 \\
\hline
Swin-S & 838G & 107M & 3x & 51.9 & 70.7 & 56.3 & 35.2 & 55.7 & 67.7 & 45.0 & 68.2 & 48.8 & 28.8 & 48.7 & 60.6 \\
\rowcolor{mygray}
\textbf{\ourmethod-Swin-S} & 838G & 107M & 3x & 52.5 & 71.3 & 57.2 & 35.6 & 56.1 & 68.0 & 45.4 & 68.9 & 49.6 & 29.1 & 49.2 & 60.6 \\
\hline
Swin-B & 981G & 145M & 3x & 51.9 & 70.5 & 56.4 & 35.4 & 55.2 & 67.4 & 45.0 & 68.1 & 48.9 & 28.9 & 48.3 & 60.4 \\
\rowcolor{mygray}
\textbf{\ourmethod-Swin-B} & 983G & 145M & 3x & 52.7 & 71.2 & 57.2 & 37.0 & 56.1 & 68.0 & 45.5 & 68.8 & 49.6 &  30.1 & 48.8 & 60.9 \\
\thickhline
\end{tabular}}
\end{center}
\vskip -0.2in
\caption{Results on COCO dataset. The FLOPs are computed over backbone, FPN and detection head with input resolution of 1280$\times$800. }
\label{tab:det2}
\vskip -0.17in
\end{table*}


\begin{table}[t]\footnotesize
\begin{center}
\setlength{\tabcolsep}{1.4mm}{
\renewcommand\arraystretch{1.05}
\begin{tabular}{l|c|ccc|ccc}
\thickhline
\multicolumn{8}{c}{\textbf{RetinaNet Object Detection on COCO (Sch. 1x)}} \\
Method & FLOPs & AP & AP$_\text{50}$ & AP$_\text{75}$ & AP$_{s}$ & AP$_{m}$ & AP$_{l}$ \\
\hline
PVT-T & 221G & 36.7 & 56.9 & 38.9 & 22.6 & 38.8 & 50.0 \\
\rowcolor{mygray}
\textbf{\ourmethod-PVT-T} & 200G & 40.1 & 61.1 & 42.2 & 25.9 & 43.3 & 54.2 \\
\hline
PVT-S & 286G &  38.7 & 59.3 & 40.8 & 21.2 & 41.6 & 54.4 \\
\rowcolor{mygray}
\textbf{\ourmethod-PVT-S} & 251G & 42.4 & 63.9 & 45.0 & 26.8 & 45.6 & 56.9 \\
\hline
PVT-M & 373G & 41.9 & 63.1 & 44.3 & 25.0 & 44.9 & 57.6 \\
\rowcolor{mygray}
\textbf{\ourmethod-PVT-M} & 338G & 43.5 & 64.7 & 46.1 & 26.3 & 47.1 & 58.5 \\
\hline
PVTV2-B3 & 379G & 45.9 & 66.8 & 49.3 & 28.6 & 49.8 & 61.4 \\
\rowcolor{mygray}
\textbf{\ourmethod-PVTv2-B3} & 343G & 46.8 & 67.7 & 50.3 & 30.5 & 51.1 & 61.6\\
\thickhline
\end{tabular}}
\end{center}
\vskip -0.2in
\caption{Results on COCO object detection with RetinaNet \cite{rtn}. The FLOPs are computed over backbone, FPN, and detection head with an input resolution of 1280$\times$800. }
\label{tab:det1}
\vskip -0.15in
\end{table}

\section{Experiments}

We conduct experiments on several datasets to verify the performance of our Slide Attention module. We show comparison results on ImageNet \cite{in1k} classification, ADE20K \cite{ade20k} semantic segmentation and COCO \cite{coco} object detection tasks. We also provide a detailed comparison with other local attention modules based on two representative model structures. In addition, ablation studies are conducted to show the effectiveness of the designs in our module. See Appendix for detailed dataset and training configurations.

\subsection{ImageNet-1K Classification}
ImageNet-1K \cite{in1k} contains 1.28M images for training and 50K images for validation. We practically implement our module on five advanced Vision Transformer models, 
and compare with various state-of-the-art models.

We show the classification results in Fig.\ref{main}. It is shown that our method achieves consistent improvements against baseline models under comparable FLOPs or parameters. For example, based on PVT, our model achieves even 0.5\% higher performance, with 60\% FLOPs. Our model based on PVTv2 and Swin Transformer also achieve comparable performance with 60\%-70\% FLOPs of competitive baselines. These results demonstrate that our module is applicable to various model structures and shows a better trade-off between computation cost and model performance.

\subsection{ADE20K Semantic Segmentation}

ADE20K \cite{ade20k} is a widely adopted benchmark for semantic segmentation with 20K training and 2K validation images. We employ our model on two representative segmentation models, SemanticFPN \cite{semfpn} and UperNet \cite{upernet}. The comparison results show that our model can be adopted on various segmentation frameworks and effectively improve the model performance on dense prediction task.

\begin{table}[t]\footnotesize
\newcommand{\tabincell}[2]{\begin{tabular}{@{}#1@{}}#2\end{tabular}}
\begin{center}
\setlength{\tabcolsep}{2.2mm}{
\renewcommand\arraystretch{1.05}
\begin{tabular}{l|c|cc|cc}
\thickhline
\multicolumn{6}{c}{\textbf{Semantic Segmentation on ADE20K}} \\
Backbone & Method & FLOPs & \#Params & mIoU & mAcc \\
\hline
PVT-T & S-FPN & 158G & 17M & 36.57 & 46.72 \\
\rowcolor{mygray}
\textbf{\ourmethod-PVT-T} & S-FPN & 136G & 16M & \textbf{38.43} & 50.05 \\
\hline
PVT-S & S-FPN & 225G & 28M & 41.95 & 53.02 \\
\rowcolor{mygray}
\textbf{\ourmethod-PVT-S} & S-FPN & 188G & 26M & \textbf{42.47} & 54.00 \\
\hline
Swin-T & UperNet & 945G & 60M & 44.51 & 55.61 \\
\rowcolor{mygray}
\textbf{\ourmethod-Swin-T} & UperNet & 946G & 60M & \textbf{45.67} & 57.13 \\
\hline
Swin-S & UperNet & 1038G & 81M & 47.64 & 58.78 \\
 \rowcolor{mygray}
\textbf{\ourmethod-Swin-S} & UperNet & 1038G & 81M & \textbf{48.46} & 60.18 \\
\thickhline
\end{tabular}}
\end{center}
\vskip -0.2in
\caption{Results of semantic segmentation. The FLOPs are computed over encoders and decoders with an input image at the resolution of 512$\times$2048. S-FPN is short for SemanticFPN \cite{semfpn} model.}
\label{tab:seg}
\vskip -0.15in
\end{table}

\begin{figure*}[t]
    \centering
    \includegraphics[width=0.95\linewidth]{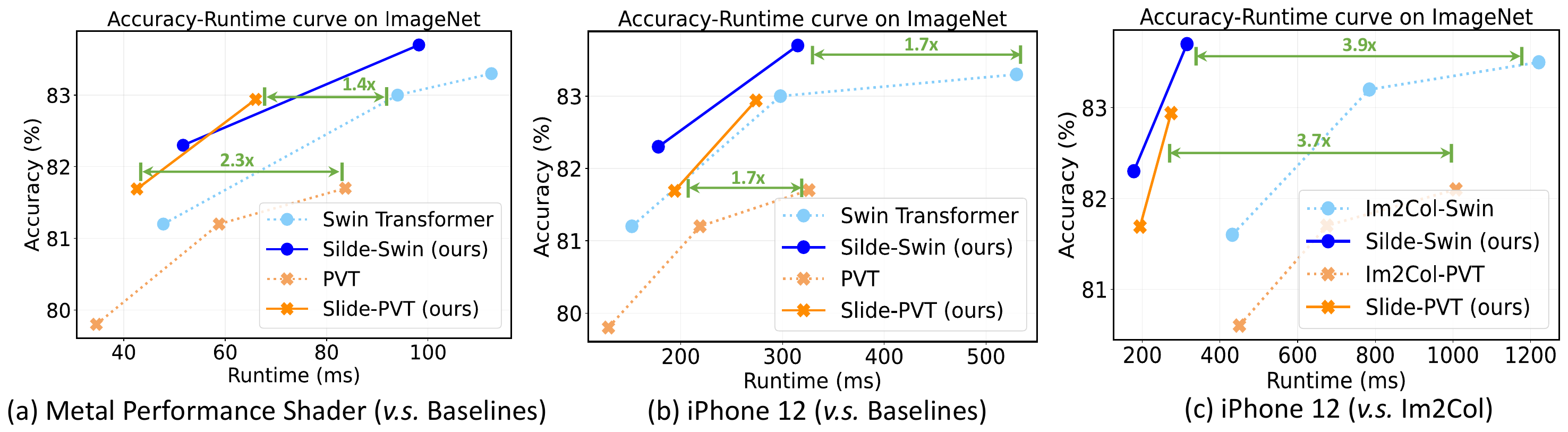}
    \vskip -0.12in
    \caption{Runtime comparison on Metal Performance Shader and iPhone 12 devices.}
    \label{fig:speed}
    \vskip -0.17in
\end{figure*}

\subsection{COCO Object Detection}

COCO \cite{coco} object detection and instance segmentation dataset has 118K training and 5K validation images. We use ImageNet pretrained model as the backbone in RetinaNet \cite{rtn}, Mask R-CNN \cite{mrcn} and Cascade Mask R-CNN \cite{cmrcn} frameworks to evaluate the effectiveness of our method.

We conduct experiments on both 1x and 3x schedules with different detection heads and show results in Tab.\ref{tab:det2} and Tab.\ref{tab:det1}. Our model shows better results under all settings. Also, our model achieves more significant improvements in detecting \textbf{small} objects (up to 4.5\% improvement), which demonstrates the effectiveness of injecting local inductive bias towards Vision Transformer backbones.

\subsection{Comparison with Other Local Attentions}
To show a fair comparison with other local attention modules, we select two representative Vision Transformer models, Swin Transformer~\cite{swin} and NAT~\cite{nat}, that are originally constructed based on window attention and local attention respectively. We adapt previous local attention approaches, including SASA~\cite{sasa}, SAN~\cite{san}, and NAT~\cite{nat} into these two models, and compare performance with ours.

As shown in Tab.\ref{tab:local_attention}, our model achieves significantly better results than Im2Col-based approach SASA. When comparing with CUDA-based approaches SAN and NAT, our model achieves higher performances (0.5\%-1.3\%) with comparable inference speed. This demonstrates the comprehensive superiority of our model on the accuracy-efficiency trade-off against other local attention approaches.

\begin{table}[t]\footnotesize
\newcommand{\tabincell}[2]{\begin{tabular}{@{}#1@{}}#2\end{tabular}}
\begin{center}
\setlength{\tabcolsep}{3.5mm}{
\renewcommand\arraystretch{1.2}
\begin{tabular}{c|cc|cc}
\thickhline
\multicolumn{5}{c}{\textbf{(a) Comparison on Swin-T Setting}} \\
Local Attention & FLOPs & \#Param & Acc. & FPS\\
\hline
SASA~\cite{sasa} & 4.5G & 29M & 81.6 & 644 \\
SAN~\cite{san} & 4.5G & 29M & 81.4 & 670 \\
NAT~\cite{nat} & 4.5G & 29M & 81.8 & 821 \\
\hline
\rowcolor{mygray}
\textbf{Ours} & 4.6G & 30M & \textbf{82.3} & 790 \\
\thickhline
\multicolumn{5}{c}{\textbf{(b) Comparison on NAT-Mini Setting}} \\
Local Attention & FLOPs & \#Param & Acc. & FPS\\
\hline
SASA~\cite{sasa} & 2.7G & 20M & 81.2 & 791\\
SAN~\cite{san} & 2.7G & 20M & 81.1 & 815\\
NAT~\cite{nat} & 2.7G & 20M & 81.8 & 1045\\
\hline
\rowcolor{mygray}
\textbf{Ours} & 2.7G & 20M & \textbf{82.4} & 998\\
\thickhline
\end{tabular}}
\end{center}
\vskip -0.22in
\caption{Comparison of different local attention modules on different model structures. We use Swin-Transformer and NAT as the basic settings. FPS is tested on a single RTX3090 GPU.}
\label{tab:local_attention}
\vskip -0.2in
\end{table}

\subsection{Inference Time}
We further investigate the practical inference time of our method under different hardware, including computation units like Metal Performance Shader (MPS) and edge devices like iPhone 12. We show comparison results with two competitive baselines in Fig.\ref{fig:speed}. We can see that our module shows significantly better trade-off between runtime and model performance on different devices, and achieves up to 2.3x speed up on advanced Vision Transformer models. For other local attention modules, due to the fact that CUDA-based approaches cannot be implemented on these devices, we only compare our method with Im2Col-based approach. As shown in Fig.\ref{fig:speed}(c), our model achieves 3.7x-3.9x speed up while maintaining higher performances.

\subsection{Ablation Study}
To further validate the effectiveness of the designs in our model, we conduct several ablation studies. As shown in Tab.\ref{tab:abl}, we can see that our slide attention module shows better performances when adopted at the early stages of Transformer models. Considering that our module has a similar design pattern with convolution, we believe this result is in accordance with the previous finding in \cite{early}, that convolutions are more useful at the early stages of Vision Transformer. We also show the effectiveness of each module in slide attention in Fig.\ref{fig1}, which contributes to better model performance or efficiency respectively.

\begin{table}[t]\footnotesize
\newcommand{\tabincell}[2]{\begin{tabular}{@{}#1@{}}#2\end{tabular}}
\begin{center}
\setlength{\tabcolsep}{1.2mm}{
\renewcommand\arraystretch{1.2}
\begin{tabular}{cccc|cc|cc}
\thickhline
\multicolumn{4}{c|}{Stages w/ Slide Attention} & \multirow{2}{*}{FLOPs} & \multirow{2}{*}{\#Param} & \multirow{2}{*}{Acc.} & \multirow{2}{*}{Diff.}\\
Stage1 & Stage2 & Stage3 & Stage4 & & & \\
\hline
\ding{51} & & & & 4.5G & 29M & 81.8 & -0.5\\
\ding{51} & \ding{51} & & & 4.6G & 29M & \textbf{82.3} & \textbf{Ours}\\
\ding{51} & \ding{51} & \ding{51} &  & 4.6G & 30M & 82.2 & -0.1\\
\ding{51} & \ding{51} & \ding{51} & \ding{51} & 4.7G & 30M & 81.3 & -1.0\\
\hline
\multicolumn{4}{c|}{Swin-T \cite{swin}} & 4.5G & 29M & 81.3 & -1.0\\
\thickhline
\end{tabular}}
\end{center}
\vskip -0.22in
\caption{Ablation study on applying slide attention on different stages. All the models are based on the Swin-Tiny structure.}
\label{tab:abl}
\vskip -0.2in
\end{table}










\section{Conclusion}

In this paper, we revisit the local attention mechanism and address its efficiency overhead by proposing a novel Slide Attention module with only common convolution operations. By substituting the inefficient Im2Col function with depthwise convolutions and equipped with a deformed shifting module, our module realizes local attention in high efficiency, flexibility, and generalizability. Extensive experiments demonstrated that our module can be widely adopted on a variety of Vision Transformers and different hardware devices while achieving a better trade-off between computation efficiency and model performance.

\section*{Acknowledgement}
This work is supported in part by the National Key R\&D Program of China (2019YFC1408703), the National Natural Science Foundation of China (62022048, 62276150), Guoqiang Institute of Tsinghua University and Beijing Academy of Artificial Intelligence. We also appreciate generous donation of computing resources by High-Flyer AI.

\section*{Supplementary Material}
\section*{A. Model Architectures}
We summarize the architectures of five Transformer models adopted in the main paper, including PVT~\cite{pvt}, PVTv2~\cite{pvtv2}, Swin Transformer~\cite{swin}, CSwin Transformer~\cite{cswin}, NAT~\cite{nat} in Tab.\ref{model_PVT}-\ref{model_nat}. For fair comparison, we only substitute the original self-attention blocks at early stages of the baseline models with our proposed \textit{Slide Attention}, while the remaining blocks, training configurations, and model structure (width and depth) are kept unchanged.

\section*{B. Dataset and Training Setup}
\noindent
\textbf{ImageNet.} ImageNet 2012 \cite{in1k} comprises 1.28 million training images and 50,000 validation images from 1000 different classes. For all baseline models, we follow the training configurations in the original paper, with adamw optimizer, 300 epoch training, and data augmentation settings follow DeiT~\cite{deit}. For CSwin-Transformer, we follow the original setting and use exponential moving average (EMA)~\cite{ema} with the same ema decay rate.

\noindent
\textbf{COCO.} COCO dataset \cite{coco} is a standard object detection benchmark and we use a subset of 80k samples as training set and 35k for validation. For all baseline models, we train the network by adamw. Backbone networks are respectively pretrained on ImageNet dataset following the same training configurations in the original paper. We follow the "1x" learning schedule to train the whole network for 12 epochs and divide the learning rate by 10 at the 8th and 11th epoch respectively. For several models, we follow the configurations in the original paper, and additionally experiment "3x" schedule with 36 epochs.
We apply standard data augmentation, that is resize, random flip and normalize. Learning rate is set at 0.01 and linear warmup is used in the first 500 iterations. We follow the "1x" learning schedule training the whole network for 12 epochs and divide the learning rate by 10 at the 8th and 11th epoch respectively. For several transformer-based models, we follow the configurations in the original paper, and test with "3x" schedule. All mAP results in the main paper are tested with input image size (3, 1333, 800).

\noindent
\textbf{ADE20K.} ADE20K \cite{ade20k} is a widely-used semantic segmentation dataset, containing 150 categories. ADE20K has 25K images, with 20K for training, 2K for validation, and another 3K for testing. For baseline models, we follow the training configurations in their original paper respectively. For Semantic FPN \cite{semfpn}, we optimize the models using AdamW with an initial learning rate of 1e-4 for 80k iterations. For UperNet \cite{upernet}, we use the AdamW optimizer with an initial learning rate of 6e-5 and a linear warmup of 1,500 iterations. Models are trained for a total of 160K iterations. We randomly resize and crop the image to 512 × 512 for training, and re-scale to have a shorter side of 512 pixels during testing.

\begin{table}[h]
\begin{center}
\setlength{\tabcolsep}{0.5mm}{
\renewcommand\arraystretch{1.2}
\begin{tabular}{l|c|ccc|ccc}
\thickhline
\multicolumn{8}{c}{\textbf{RetinaNet Object Detection on COCO (Sch. 1x)}} \\
Method & FLOPs & AP & AP$_\text{50}$ & AP$_\text{75}$ & AP$_{s}$ & AP$_{m}$ & AP$_{l}$ \\
\hline
PVT-T & 221G & 36.7 & 56.9 & 38.9 & 22.6 & 38.8 & 50.0 \\
\rowcolor{mygray}
\textbf{\ourmethod-PVT-T} & 200G & 40.1 & 61.1 & 42.2 & 25.9 & 43.3 & 54.2 \\
\hline
PVT-S & 286G &  38.7 & 59.3 & 40.8 & 21.2 & 41.6 & 54.4 \\
\rowcolor{mygray}
\textbf{\ourmethod-PVT-S} & 251G & 42.4 & 63.9 & 45.0 & 26.8 & 45.6 & 56.9 \\
\hline
PVT-M & 373G & 41.9 & 63.1 & 44.3 & 25.0 & 44.9 & 57.6 \\
\rowcolor{mygray}
\textbf{\ourmethod-PVT-M} & 338G & 43.5 & 64.7 & 46.1 & 26.3 & 47.1 & 58.5 \\
\hline
PVTV2-B0 & 178G & 37.2 & 57.2 & 39.5 & 23.1 & 40.4 & 49.7 \\
\rowcolor{mygray}
\textbf{\ourmethod-PVTv2-B0} & 167G & 37.6 & 57.9 & 40.1 & 22.9 & 40.4 & 50.2 \\
\hline
PVTV2-B1 & 225G & 41.2 & 61.9 & 43.9 & 25.4 & 44.5 & 54.3 \\
\rowcolor{mygray}
\textbf{\ourmethod-PVTv2-B1} & 204G & 41.5 & 62.3 & 44.0 & 26.0 & 44.8 & 54.9 \\
\hline
PVTV2-B2 & 291G & 44.6 & 65.6 & 47.6 & 27.4 & 48.8 & 58.6 \\
\rowcolor{mygray}
\textbf{\ourmethod-PVTv2-B2} & 255G & 45.0 & 66.2 & 48.4 & 28.8 & 48.8 & 59.7 \\
\hline
PVTV2-B3 & 379G & 45.9 & 66.8 & 49.3 & 28.6 & 49.8 & 61.4 \\
\rowcolor{mygray}
\textbf{\ourmethod-PVTv2-B3} & 343G & 46.8 & 67.7 & 50.3 & 30.5 & 51.1 & 61.6\\
\thickhline
\end{tabular}}
\end{center}
\caption{Results on COCO object detection with RetinaNet \cite{rtn}. The FLOPs are computed over backbone, FPN, and detection head with an input resolution of 1280$\times$800. }
\label{tab:det1}
\end{table}

\begin{table}[h]
\newcommand{\tabincell}[2]{\begin{tabular}{@{}#1@{}}#2\end{tabular}}
\begin{center}
\setlength{\tabcolsep}{3.0mm}{
\renewcommand\arraystretch{1.2}
\begin{tabular}{l|cc|l}
\thickhline
\textbf{Method} & \textbf{Params} & \textbf{Flops} & \textbf{Top-1}\\
\hline
PVT-T~\cite{pvt}  & 13.2M & 1.9G & 75.1\\
\rowcolor{mygray}
\textbf{Slide-PVT-T} & 12.2M & 2.0G & \textbf{78.0\,{\scriptsize (+2.9)}}\\
PVT-S & 24.5M & 3.8G & 79.8\\
\rowcolor{mygray}
\textbf{Slide-PVT-S} & 22.7M & 4.0G & \textbf{81.7\,{\scriptsize (+1.9)}}\\
PVT-M & 44.2M & 6.7G & 81.2\\
\rowcolor{mygray}
\textbf{Slide-PVT-M} & 42.5M & 9.8G & \textbf{82.9\,{\scriptsize (+1.7)}}\\
PVT-L & 61.4M & 6.7G & 81.7\\
\rowcolor{mygray}
\textbf{Slide-PVT-L} & 59.8M & 9.8G & \textbf{83.9\,{\scriptsize (+2.2)}}\\
\hline
PVTv2-B0~\cite{pvtv2}  & 3.4M & 0.6G & 70.5\\
\rowcolor{mygray}
\textbf{Slide-PVTv2-B0} & 3.3M & 0.6G & \textbf{71.4\,{\scriptsize (+0.9)}}\\
PVTv2-B1 & 13.1M & 2.1G & 78.7\\
\rowcolor{mygray}
\textbf{Slide-PVTv2-B1} & 13.0M & 2.2G & \textbf{79.5\,{\scriptsize (+0.7)}}\\
PVTv2-B2  & 25.4M & 4.0G & 82.0\\
\rowcolor{mygray}
\textbf{Slide-PVTv2-B2} & 22.8M & 4.2G & \textbf{82.7\,{\scriptsize (+0.7)}}\\
PVTv2-B3  & 45.2M & 6.9G & 83.2\\
\rowcolor{mygray}
\textbf{Slide-PVTv2-B3} & 42.5M & 7.1G & \textbf{83.8\,{\scriptsize (+0.6)}}\\
PVTv2-B4  & 62.6 M & 10.1G & 83.6\\
\rowcolor{mygray}
\textbf{Slide-PVTv2-B4} & 59.8M & 10.3G & \textbf{84.2\,{\scriptsize (+0.6)}}\\
PVTv2-B5  & 82.0M & 11.8G & 83.8\\
\rowcolor{mygray}
\textbf{Slide-PVTv2-B5} & 78.9M & 12.1G & \textbf{84.3\,{\scriptsize (+0.5)}}\\
\hline
Swin-T~\cite{swin}  & 29M & 4.5G & 81.3\\
\rowcolor{mygray}
\textbf{Slide-Swin-T} & 29M & 4.6G & \textbf{82.3\,{\scriptsize (+1.0)}}\\
Swin-S & 50M & 8.7G & 83.0\\
\rowcolor{mygray}
\textbf{Slide-Swin-S} & 51M & 8.9G & \textbf{83.7\,{\scriptsize (+0.7)}}\\
Swin-B & 88M & 15.4G & 83.5\\
\rowcolor{mygray}
\textbf{Slide-Swin-B} & 89M & 15.5G & \textbf{84.2\,{\scriptsize (+0.7)}}\\
\hline
CSwin-T~\cite{cswin} & 23M & 4.3G & 82.8\\
\rowcolor{mygray}
\textbf{Slide-CSwin-T} & 23M & 4.3G & \textbf{83.2\,{\scriptsize (+0.4)}}\\
CSwin-S & 35M & 6.9G & 83.6\\
\rowcolor{mygray}
\textbf{Slide-CSwin-S} & 35M & 6.9G & \textbf{84.0\,{\scriptsize (+0.4)}}\\
CSwin-B & 78M & 15.0G & 84.2\\
\rowcolor{mygray}
\textbf{Slide-CSwin-B} & 78M & 15.0G & \textbf{84.7\,{\scriptsize (+0.5)}}\\
\hline
NAT-M~\cite{nat} & 20M & 2.7G & 81.8\\
\rowcolor{mygray}
\textbf{Slide-NAT-M} & 20M & 2.7G & \textbf{82.4\,{\scriptsize (+0.6)}}\\
NAT-T & 28M & 4.3G & 83.2\\
\rowcolor{mygray}
\textbf{Slide-NAT-T} & 28M & 4.3G & \textbf{83.6\,{\scriptsize (+0.4)}}\\
NAT-S & 51M & 7.8G & 83.7\\
\rowcolor{mygray}
\textbf{Slide-NAT-S} & 51M & 7.8G & \textbf{84.3\,{\scriptsize (+0.6)}}\\
\thickhline
\end{tabular}}
\end{center}
\vskip -0.1in
\caption{Comparisons of slide attention with other vision transformer backbones on FLOPs, parameters, accuracy on the ImageNet-1K classification task.}
\label{tab:cls}
\end{table}

\begin{table}[h]
\newcommand{\tabincell}[2]{\begin{tabular}{@{}#1@{}}#2\end{tabular}}
\begin{center}
\setlength{\tabcolsep}{1.0mm}{
\renewcommand\arraystretch{1.2}
\begin{tabular}{l|c|cc|cc}
\thickhline
\multicolumn{6}{c}{\textbf{Semantic Segmentation on ADE20K}} \\
Backbone & Method & FLOPs & \#Params & mIoU & mAcc \\
\hline
PVT-T & S-FPN & 158G & 17M & 36.57 & 46.72 \\
\rowcolor{mygray}
\textbf{\ourmethod-PVT-T} & S-FPN & 136G & 16M & \textbf{38.43} & 50.05 \\
\hline
PVT-S & S-FPN & 225G & 28M & 41.95 & 53.02 \\
\rowcolor{mygray}
\textbf{\ourmethod-PVT-S} & S-FPN & 188G & 26M & \textbf{42.47} & 54.00 \\
\hline
PVT-M & S-FPN & 315G & 48M & 42.91 & 53.80 \\
\rowcolor{mygray}
\textbf{\ourmethod-PVT-M} & S-FPN & 278G & 46M & \textbf{43.97} & 55.58 \\
\hline
Swin-T & UperNet & 945G & 60M & 44.51 & 55.61 \\
\rowcolor{mygray}
\textbf{\ourmethod-Swin-T} & UperNet & 946G & 60M & \textbf{45.67} & 57.13 \\
\hline
Swin-S & UperNet & 1038G & 81M & 47.64 & 58.78 \\
 \rowcolor{mygray}
\textbf{\ourmethod-Swin-S} & UperNet & 1038G & 81M & \textbf{48.46} & 60.18 \\
\hline
Swin-B & UperNet & 1188G & 121M & 48.13 & 59.13 \\
 \rowcolor{mygray}
\textbf{\ourmethod-Swin-B} & UperNet & 1188G & 121M & \textbf{48.58} & 60.26 \\
\thickhline
\end{tabular}}
\end{center}
\vskip -0.1in
\caption{Results of semantic segmentation. The FLOPs are computed over encoders and decoders with an input image at the resolution of 512$\times$2048. S-FPN is short for SemanticFPN \cite{semfpn} model.}
\label{tab:seg}
\end{table}
\vskip -0.1in

\section*{C. Additional Downstream Experiments}
We also provide additional experiment results on semantic segmentation and object detection, and show in Tab.\ref{tab:det1}, Tab.\ref{tab:seg} and Tab.\ref{tab:det2}. Similar pattern can be observed: (1) Our module achieves consistent improvements over baseline models; (2) The improvements on small objects are more significant, indicating the effectiveness of local inductive bias in our slide attention module. We also test the runtime for segmentation and detection on a RTX3090 GPU, and show the results below. Similar to classification, our method achieves a better trade-off than baselines.

\begin{figure}[h]
    \centering
    \includegraphics[width=0.95\linewidth]{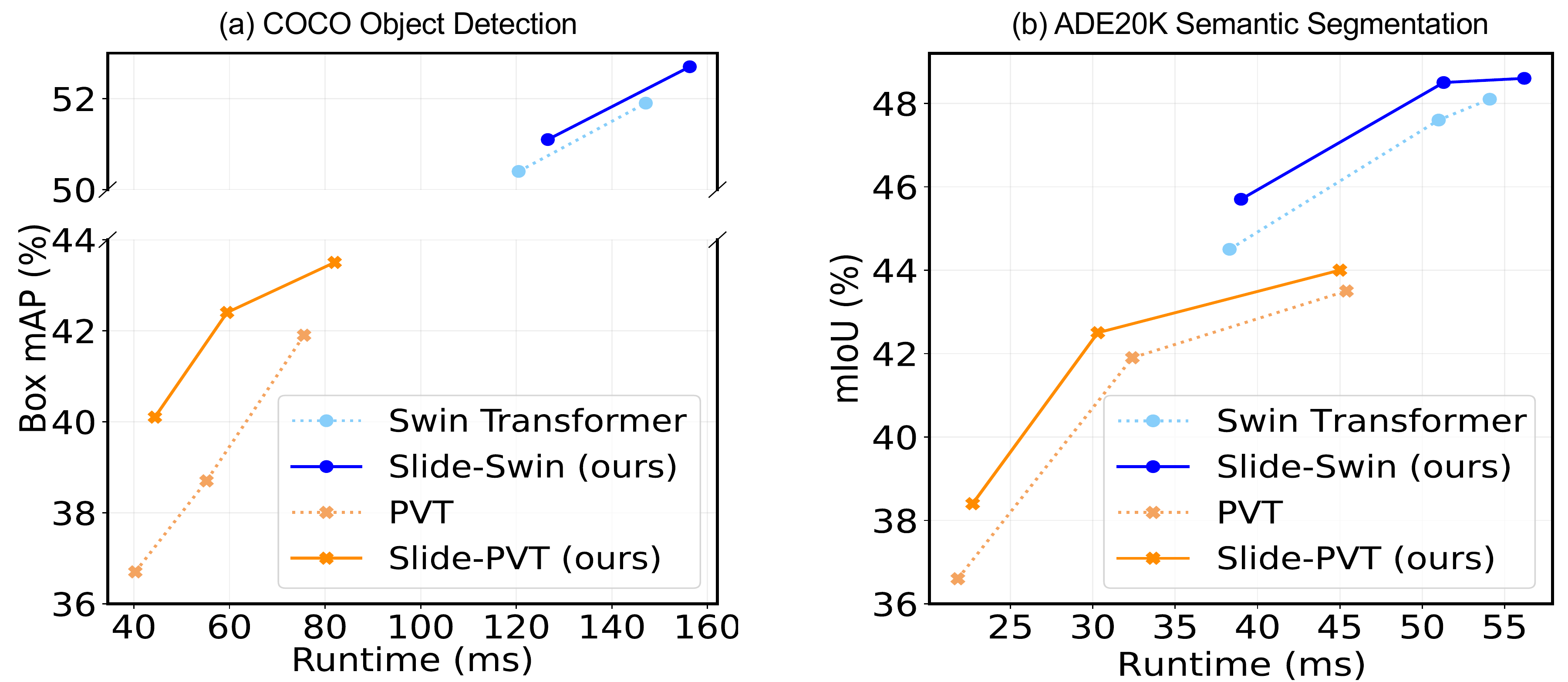}
    \vskip -0.1in
    \caption{Runtime for object detection and segmentation.}
    \label{fig:fig1}
    \vskip -0.15in
\end{figure}

\section*{D. Full Classification Results}
Due to the page limit, we only present representative ImageNet classification results in Figure 5 of main paper. Here, we show the full classification results when adapting our module on all model sizes of baseline models in Tab.\ref{tab:cls}.

\begin{figure*}[h]
    \centering
    \includegraphics[width=0.7\linewidth]{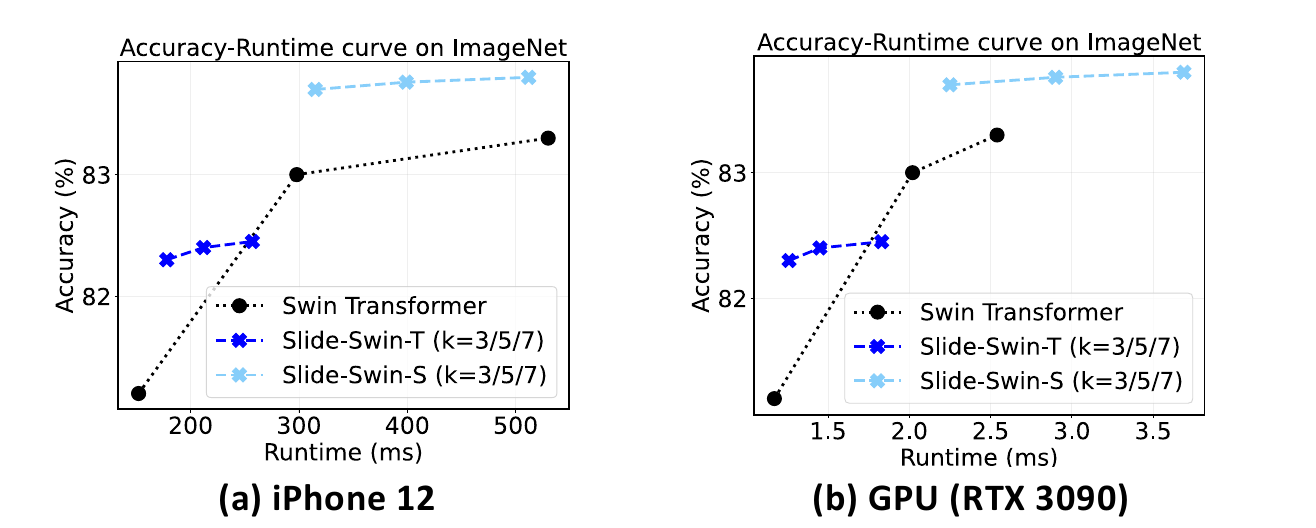}
    \vskip -0.1in
    \caption{Ablation study on the window size $k$. We show runtime comparison on iPhone 12 and RTX 3090 GPU.}
    \label{fig:speed}
\end{figure*}
\vskip -0.1in

\begin{table*}[t]
\begin{center}
\setlength{\tabcolsep}{1.0mm}{
\renewcommand\arraystretch{1.2}
\begin{tabular}{l|c|c|c|ccc|ccc|ccc|ccc}
\thickhline
\multicolumn{16}{c}{\textbf{(a) Mask R-CNN Object Detection \& Instance Segmentation on COCO}} \\
Method & FLOPs & \#Param & Schedule & AP$^b$ & AP$^b_\text{50}$ & AP$^b_\text{75}$ & AP$^b_s$ & AP$^b_m$ & AP$^b_l$ & AP$^m$ & AP$^m_\text{50}$ & AP$^m_\text{75}$ & AP$^m_s$ & AP$^m_m$ & AP$^m_l$ \\
\hline
PVT-T & 240G & 33M & 1x & 36.7 & 59.2 & 39.3 & 21.6 & 39.2 & 49.0 & 35.1 & 56.7 & 37.3 & 19.5 & 37.4 & 48.5 \\
\rowcolor{mygray}
\textbf{\ourmethod-PVT-T} & 219G & 32M & 1x & 40.4 & 63.4 & 43.8 & 25.3 & 42.8 & 53.0 & 38.1 & 60.4 & 41.0 & 20.0 & 40.1 & 55.2 \\
\hline
PVT-S & 305G & 44M & 1x & 40.4 & 62.9 & 43.8 & 22.9 & 43.0 & 55.4 & 37.8 & 60.1 & 40.3 & 20.4 & 40.3 & 53.6 \\
\rowcolor{mygray}
\textbf{\ourmethod-PVT-S} & 269G & 42M & 1x & 42.8 & 65.9 & 46.7 & 26.6 & 45.5 & 57.3 & 40.1  & 63.1 & 43.1 & 20.3 & 42.4 & 59.0 \\
\hline
PVT-M & 392G & 64M & 1x & 42.0 & 64.4 & 45.6 & 24.4 & 44.9 & 57.9 & 39.0 & 61.6 & 42.1 & 21.3 & 42.0 & 55.2 \\
\rowcolor{mygray}
\textbf{\ourmethod-PVT-M} & 357G & 62M & 1x & 44.4 & 66.9 & 48.6 & 28.9 & 47.0 & 59.4 & 0.408  & 63.9 & 43.8 & 25.0 & 43.5 & 55.9 \\
\hline
PVTv2-B0 & 196G & 23M & 1x & 38.2 & 60.5 & 40.7 & 22.9 & 40.9 & 49.6 & 36.2 & 57.8 & 38.6 & 18.0 & 38.4 & 51.9 \\
\rowcolor{mygray}
\textbf{\ourmethod-PVTv2-B0} & 185G & 23M & 1x & 38.8 & 60.9 & 41.9 & 23.7 & 41.5 & 50.7 & 36.4 & 58.0 & 38.8 & 20.6 & 39.1 & 49.5 \\
\hline
PVTv2-B1 & 244G & 34M & 1x & 41.8 & 64.3 & 45.9 & 26.4 & 44.9 & 54.3 & 38.8 & 61.2 & 41.6 & 20.2 & 41.3 & 56.1 \\
\rowcolor{mygray}
\textbf{\ourmethod-PVTv2-B1} & 222G & 33M & 1x & 42.6 & 65.3 & 46.8 & 27.4 & 45.6 & 55.7 & 39.7 & 62.6 & 42.6 & 24.1 & 42.9 & 53.7 \\
\hline
PVTv2-B2 & 309G & 45M & 1x & 45.3 & 67.1 & 49.6 & 28.8 & 48.4 & 59.5 & 41.2 & 64.2 & 44.4 & 22.0 & 43.7 & 59.4 \\
\rowcolor{mygray}
\textbf{\ourmethod-PVTv2-B2} & 274G & 43M & 1x & 46.0 & 68.2 & 50.3 & 28.8 & 49.4 & 61.0 & 41.9 & 65.1 & 45.4 & 24.6 & 45.2 & 57.2 \\
\hline
PVTv2-B3 & 397G & 65M & 1x & 47.0 & 68.1 & 51.7 & 30.2 & 50.4 & 62.4 & 42.5 & 65.7 & 45.7 & 23.2 & 45.3 &  61.5 \\
\rowcolor{mygray}
\textbf{\ourmethod-PVTv2-B3} & 362G & 63M & 1x & 47.8 & 69.5 & 52.6 & 30.2 & 51.3 & 62.8 & 43.2 & 66.5 & 46.6 & 26.1 & 46.3 & 58.7 \\
\hline
Swin-T & 267G & 48M & 1x & 43.7 & 66.6 & 47.7 & 28.5 & 47.0 & 57.3 & 39.8 & 63.3 & 42.7 & 24.2 & 43.1 & 54.6 \\
\rowcolor{mygray}
\textbf{\ourmethod-Swin-T} & 268G & 48M & 1x & 44.3 & 67.2 & 48.5 & 28.9 & 47.8 & 57.0 & 40.3 & 63.9 & 43.0 & 24.3 & 44.0 & 54.5 \\
Swin-T & 267G & 48M & 3x & 46.0 & 68.1 & 50.3 & 31.2 & 49.2 & 60.1 & 41.6 & 65.1 & 44.9 & 25.9 & 45.1 & 56.9 \\
\rowcolor{mygray}
\textbf{\ourmethod-Swin-T} & 268G & 48M & 3x & 46.8 & 69.0 & 51.6 & 31.7 & 50.4 & 60.1 & 42.3 & 66.0 & 45.8 & 23.5 & 45.8 & 60.8 \\
\thickhline
\multicolumn{16}{c}{\textbf{(b) Cascade Mask R-CNN Object Detection \& Instance Segmentation on COCO}} \\
Method & FLOPs & \#Param & Schedule & AP$^b$ & AP$^b_\text{50}$ & AP$^b_\text{75}$ & AP$^b_s$ & AP$^b_m$ & AP$^b_l$ & AP$^m$ & AP$^m_\text{50}$ & AP$^m_\text{75}$ & AP$^m_s$ & AP$^m_m$ & AP$^m_l$ \\
\hline
Swin-T & 745G & 86M & 1x & 48.1 & 67.1 & 52.2 & 30.4 & 51.5 & 63.1 & 41.7 & 64.4 & 45.0 & 24.0 & 45.2 & 56.9 \\
\rowcolor{mygray}
\textbf{\ourmethod-Swin-T} & 747G & 86M & 1x & 48.6 & 67.7 & 52.7 & 32.1 & 52.2 & 63.5 & 41.9 & 65.0 & 45.2 & 23.2 & 45.3 & 60.9 \\
\hline
Swin-T & 745G & 86M & 3x & 50.4 & 69.2 & 54.7 & 33.8 & 54.1 & 65.2 & 43.7 & 66.6 & 47.3 & 27.3 & 47.5 & 59.0 \\
\rowcolor{mygray}
\textbf{\ourmethod-Swin-T} & 747G & 86M & 3x & 51.1 & 69.8 & 55.4 & 35.2 & 54.4 & 65.8 & 44.3 & 67.4 & 48.0 & 28.0 & 48.0 & 59.2 \\
\hline
Swin-S & 838G & 107M & 3x & 51.9 & 70.7 & 56.3 & 35.2 & 55.7 & 67.7 & 45.0 & 68.2 & 48.8 & 28.8 & 48.7 & 60.6 \\
\rowcolor{mygray}
\textbf{\ourmethod-Swin-S} & 838G & 107M & 3x & 52.5 & 71.3 & 57.2 & 35.6 & 56.1 & 68.0 & 45.4 & 68.9 & 49.6 & 29.1 & 49.2 & 60.6 \\
\hline
Swin-B & 981G & 145M & 3x & 51.9 & 70.5 & 56.4 & 35.4 & 55.2 & 67.4 & 45.0 & 68.1 & 48.9 & 28.9 & 48.3 & 60.4 \\
\rowcolor{mygray}
\textbf{\ourmethod-Swin-B} & 983G & 145M & 3x & 52.7 & 71.2 & 57.2 & 37.0 & 56.1 & 68.0 & 45.5 & 68.8 & 49.6 &  30.1 & 48.8 & 60.9 \\
\thickhline
\end{tabular}}
\end{center}
\vskip -0.2in
\caption{Results on COCO dataset. The FLOPs are computed over backbone, FPN and detection head with input resolution of 1280$\times$800. }
\label{tab:det2}
\end{table*}

\section*{E. Ablation Study on the Window Size}
We also conduct ablation studies on the window size of our local attention, and show runtime-performance comparison on both iPhone 12 and RTX3090 GPU. Our experiments are all based on Swin-Transformer-T and Swin-Transformer-S, and only the self-attention modules in the first two stages are substituted with slide attention. It can be observed that increasing the window size brings marginal improvements on the model performance while resulting in huge increase on the inference time. Similar results are also observed in other works like \cite{qna}. Therefore, we consider using window size $k=3$ for all the results shown in the main paper. We see the ineffectiveness of increasing window size as a future direction and where a better performance-efficiency trade-off will be more valuable.

\section*{F. Limitations}
As we have stated above, the increasing of window size in slide attention only results in marginal improvements. We believe a better performance-efficiency trade-off with larger window size worth further investigation, and we see this as an important future direction.

\begin{table*}[h]\small
    \centering
    \setlength{\tabcolsep}{1mm}{
    \renewcommand\arraystretch{1.2}
    \begin{tabular}{c|c|c|c|c|c|c|c}
    \thickhline
    \multirow{2}*{stage} & \multirow{2}*{output} & \multicolumn{2}{c|}{Slide-PVT-T} & \multicolumn{2}{c|}{Slide-PVT-S} & \multicolumn{2}{c}{Slide-PVT-M}\\
    \cline{3-8}
    & & \textbf{Slide Attention} & PVT Block & \textbf{Slide Attention} & PVT Block& \textbf{Slide Attention} & PVT Block\\
    \hline
    \multirow{4}*{res1} & \multirow{4}*{$56\times 56$} & \multicolumn{6}{c}{Conv1×1, stride=4, 64, LN}\\
    \cline{3-8}
    && $\left[\!\!\! \begin{array}{c} {\rm \ win}  \ 3\!\times\! 3\\{\rm dim} \ 64 \\ {\rm head} \ 1\end{array} \!\!\! \right ] \!\!\times\! 2$ & None & $\left[\!\!\! \begin{array}{c} {\rm \ win}  \ 3\!\times\! 3\\{\rm dim} \ 64 \\ {\rm head} \ 1\end{array} \!\!\! \right ] \!\!\times\! 3$ & None & $\left[\!\!\! \begin{array}{c} {\rm \ win}  \ 3\!\times\! 3\\{\rm dim} \ 64 \\ {\rm head} \ 1\end{array} \!\!\! \right ] \!\!\times\! 3$ & None\\
    \hline
    \multirow{4}*{res2} & \multirow{4}*{$28\times 28$} & \multicolumn{6}{c}{Conv1×1, stride=2, 128, LN}\\
    \cline{3-8}
    && $\left[\!\!\! \begin{array}{c} {\rm \ win}  \ 3\!\times\! 3\\{\rm dim} \ 128 \\ {\rm head} \ 2\end{array} \!\!\! \right ] \!\!\times\! 2$ & None & $\left[\!\!\! \begin{array}{c} {\rm \ win}  \ 3\!\times\! 3\\{\rm dim} \ 128 \\ {\rm head} \ 2\end{array} \!\!\! \right ] \!\!\times\! 3$ & None & $\left[\!\!\! \begin{array}{c} {\rm \ win}  \ 3\!\times\! 3\\{\rm dim} \ 128 \\ {\rm head} \ 2\end{array} \!\!\! \right ] \!\!\times\! 3$ & None\\
    \hline
    \multirow{4}*{res3} & \multirow{4}*{$14\times 14$} & \multicolumn{6}{c}{Conv1×1, stride=2, 320, LN}\\
    \cline{3-8}
    && None & $\left[\!\!\! \begin{array}{c} {\rm \ win}  \ 7\!\times\! 7\\{\rm dim} \ 256 \\ {\rm head} \ 5\end{array} \!\!\! \right ] \!\!\times\! 2$ & None & $\left[\!\!\! \begin{array}{c} {\rm \ win}  \ 7\!\times\! 7\\{\rm dim} \ 256 \\ {\rm head} \ 5\end{array} \!\!\! \right ] \!\!\times\! 6$ & None & $\left[\!\!\! \begin{array}{c} {\rm \ win}  \ 7\!\times\! 7\\{\rm dim} \ 256 \\ {\rm head} \ 5\end{array} \!\!\! \right ] \!\!\times\! 18$\\
    \hline
    \multirow{4}*{res4} & \multirow{4}*{$7\times 7$} & \multicolumn{6}{c}{Conv1×1, stride=2, 512, LN}\\
    \cline{3-8}
    & & None& $\left[\!\!\! \begin{array}{c} {\rm \ win}  \ 7\!\times\! 7\\{\rm dim} \ 512 \\ {\rm head} \ 8\end{array} \!\!\! \right ] \!\!\times\! 2$ & None & $\left[\!\!\! \begin{array}{c} {\rm \ win}  \ 7\!\times\! 7\\{\rm dim} \ 512 \\ {\rm head} \ 8\end{array} \!\!\! \right ] \!\!\times\! 3$ & None & $\left[\!\!\! \begin{array}{c} {\rm \ win}  \ 7\!\times\! 7\\{\rm dim} \ 512 \\ {\rm head} \ 8\end{array} \!\!\! \right ] \!\!\times\! 3$\\
    \thickhline
    \end{tabular}}
    \caption{Architectures of Slide-PVT models.}
    \label{model_PVT}
\end{table*}

\begin{table*}\small
    \centering
    \setlength{\tabcolsep}{1mm}{
    \renewcommand\arraystretch{1.15}
    \begin{tabular}{c|c|c|c|c|c|c|c}
    \thickhline
    \multirow{2}*{stage} & \multirow{2}*{output} & \multicolumn{2}{c|}{Slide-PVTv2-B0} & \multicolumn{2}{c|}{Slide-PVTv2-B1} & \multicolumn{2}{c}{Slide-PVTv2-B2}\\
    \cline{3-8}
    & & \textbf{Slide Attention} & PVTv2 Block & \textbf{Slide Attention} & PVTv2 Block& \textbf{Slide Attention} & PVTv2 Block\\
    \hline
    \multirow{4}*{res1} & \multirow{4}*{$56\times 56$} & \multicolumn{2}{c|}{Conv4×4, stride=4, 32, LN} & \multicolumn{4}{c}{Conv4×4, stride=4, 64, LN}\\
    \cline{3-8}
    && $\left[\!\!\! \begin{array}{c} {\rm \ win}  \ 3\!\times\! 3\\{\rm dim} \ 32 \\ {\rm head} \ 1\end{array} \!\!\! \right ] \!\!\times\! 2$ & None & $\left[\!\!\! \begin{array}{c} {\rm \ win}  \ 3\!\times\! 3\\{\rm dim} \ 64 \\ {\rm head} \ 1\end{array} \!\!\! \right ] \!\!\times\! 2$ & None & $\left[\!\!\! \begin{array}{c} {\rm \ win}  \ 3\!\times\! 3\\{\rm dim} \ 64 \\ {\rm head} \ 1\end{array} \!\!\! \right ] \!\!\times\! 3$ & None\\
    \hline
    \multirow{4}*{res2} & \multirow{4}*{$28\times 28$} & \multicolumn{2}{c|}{Conv1×1, stride=2, 64, LN} & \multicolumn{4}{c}{Conv1×1, stride=2, 128, LN}\\
    \cline{3-8}
    && $\left[\!\!\! \begin{array}{c} {\rm \ win}  \ 3\!\times\! 3\\{\rm dim} \ 64 \\ {\rm head} \ 2\end{array} \!\!\! \right ] \!\!\times\! 2$ & None & $\left[\!\!\! \begin{array}{c} {\rm \ win}  \ 3\!\times\! 3\\{\rm dim} \ 128 \\ {\rm head} \ 2\end{array} \!\!\! \right ] \!\!\times\! 2$ & None & $\left[\!\!\! \begin{array}{c} {\rm \ win}  \ 3\!\times\! 3\\{\rm dim} \ 128 \\ {\rm head} \ 2\end{array} \!\!\! \right ] \!\!\times\! 3$ & None\\
    \hline
    \multirow{4}*{res3} & \multirow{4}*{$14\times 14$}  & \multicolumn{2}{c|}{Conv2×2, stride=2, 160, LN} & \multicolumn{4}{c}{Conv2×2, stride=2, 320, LN}\\
    \cline{3-8}
    && None & $\left[\!\!\! \begin{array}{c} {\rm \ win}  \ 7\!\times\! 7\\{\rm dim} \ 160 \\ {\rm head} \ 5\end{array} \!\!\! \right ] \!\!\times\! 2$ & None & $\left[\!\!\! \begin{array}{c} {\rm \ win}  \ 7\!\times\! 7\\{\rm dim} \ 320 \\ {\rm head} \ 5\end{array} \!\!\! \right ] \!\!\times\! 2$ & $\left[\!\!\! \begin{array}{c} {\rm \ win}  \ 3\!\times\! 3\\{\rm dim} \ 320 \\ {\rm head} \ 5\end{array} \!\!\! \right ] \!\!\times\! 2$ & $\left[\!\!\! \begin{array}{c} {\rm \ win}  \ 7\!\times\! 7\\{\rm dim} \ 320 \\ {\rm head} \ 5\end{array} \!\!\! \right ] \!\!\times\! 4$\\
    \hline
    \multirow{4}*{res4} & \multirow{4}*{$7\times 7$}  & \multicolumn{2}{c|}{Conv2×2, stride=2, 256, LN} & \multicolumn{4}{c}{Conv2×2, stride=2, 512, LN}\\
    \cline{3-8}
    & & None& $\left[\!\!\! \begin{array}{c} {\rm \ win}  \ 7\!\times\! 7\\{\rm dim} \ 512 \\ {\rm head} \ 8\end{array} \!\!\! \right ] \!\!\times\! 2$ & None & $\left[\!\!\! \begin{array}{c} {\rm \ win}  \ 7\!\times\! 7\\{\rm dim} \ 512 \\ {\rm head} \ 8\end{array} \!\!\! \right ] \!\!\times\! 2$ & None & $\left[\!\!\! \begin{array}{c} {\rm \ win}  \ 7\!\times\! 7\\{\rm dim} \ 512 \\ {\rm head} \ 8\end{array} \!\!\! \right ] \!\!\times\! 3$\\
    \thickhline
    \end{tabular}}
    \caption{Architectures of Slide-PVTv2 models (Part1).}
    \label{model_PVTv2-1}
\end{table*}

\begin{table*}\small
    \centering
    \setlength{\tabcolsep}{1mm}{
    \renewcommand\arraystretch{1.15}
    \begin{tabular}{c|c|c|c|c|c|c|c}
    \thickhline
    \multirow{2}*{stage} & \multirow{2}*{output} & \multicolumn{2}{c|}{Slide-PVTv2-B3} & \multicolumn{2}{c|}{Slide-PVTv2-B4} & \multicolumn{2}{c}{Slide-PVTv2-B5}\\
    \cline{3-8}
    & & \textbf{Slide Attention} & PVTv2 Block & \textbf{Slide Attention} & PVTv2 Block& \textbf{Slide Attention} & PVTv2 Block\\
    \hline
    \multirow{4}*{res1} & \multirow{4}*{$56\times 56$} & \multicolumn{6}{c}{Conv4×4, stride=4, 64, LN}\\
    \cline{3-8}
    && $\left[\!\!\! \begin{array}{c} {\rm \ win}  \ 3\!\times\! 3\\{\rm dim} \ 64 \\ {\rm head} \ 1\end{array} \!\!\! \right ] \!\!\times\! 3$ & None & $\left[\!\!\! \begin{array}{c} {\rm \ win}  \ 3\!\times\! 3\\{\rm dim} \ 64 \\ {\rm head} \ 1\end{array} \!\!\! \right ] \!\!\times\! 2$ & None & $\left[\!\!\! \begin{array}{c} {\rm \ win}  \ 3\!\times\! 3\\{\rm dim} \ 64 \\ {\rm head} \ 1\end{array} \!\!\! \right ] \!\!\times\! 3$ & None\\
    \hline
    \multirow{4}*{res2} & \multirow{4}*{$28\times 28$} & \multicolumn{6}{c}{Conv2×2, stride=2, 128, LN}\\
    \cline{3-8}
    && $\left[\!\!\! \begin{array}{c} {\rm \ win}  \ 3\!\times\! 3\\{\rm dim} \ 128 \\ {\rm head} \ 2\end{array} \!\!\! \right ] \!\!\times\! 3$ & None & $\left[\!\!\! \begin{array}{c} {\rm \ win}  \ 3\!\times\! 3\\{\rm dim} \ 128 \\ {\rm head} \ 2\end{array} \!\!\! \right ] \!\!\times\! 8$ & None & $\left[\!\!\! \begin{array}{c} {\rm \ win}  \ 3\!\times\! 3\\{\rm dim} \ 128 \\ {\rm head} \ 2\end{array} \!\!\! \right ] \!\!\times\! 6$ & None\\
    \hline
    \multirow{4}*{res3} & \multirow{4}*{$14\times 14$}  & \multicolumn{6}{c}{Conv2×2, stride=2, 320, LN}\\
    \cline{3-8}
    && $\left[\!\!\! \begin{array}{c} {\rm \ win}  \ 3\!\times\! 3\\{\rm dim} \ 320 \\ {\rm head} \ 5\end{array} \!\!\! \right ] \!\!\times\! 10$ & $\left[\!\!\! \begin{array}{c} {\rm \ win}  \ 7\!\times\! 7\\{\rm dim} \ 320 \\ {\rm head} \ 5\end{array} \!\!\! \right ] \!\!\times\! 8$ & $\left[\!\!\! \begin{array}{c} {\rm \ win}  \ 3\!\times\! 3\\{\rm dim} \ 320 \\ {\rm head} \ 5\end{array} \!\!\! \right ] \!\!\times\! 15$ & $\left[\!\!\! \begin{array}{c} {\rm \ win}  \ 7\!\times\! 7\\{\rm dim} \ 320 \\ {\rm head} \ 5\end{array} \!\!\! \right ] \!\!\times\! 12$ & $\left[\!\!\! \begin{array}{c} {\rm \ win}  \ 3\!\times\! 3\\{\rm dim} \ 320 \\ {\rm head} \ 5\end{array} \!\!\! \right ] \!\!\times\! 20$ & $\left[\!\!\! \begin{array}{c} {\rm \ win}  \ 7\!\times\! 7\\{\rm dim} \ 320 \\ {\rm head} \ 5\end{array} \!\!\! \right ] \!\!\times\! 20$\\
    \hline
    \multirow{4}*{res4} & \multirow{4}*{$7\times 7$}  & \multicolumn{6}{c}{Conv1×1, stride=2, 512, LN}\\
    \cline{3-8}
    & & None& $\left[\!\!\! \begin{array}{c} {\rm \ win}  \ 7\!\times\! 7\\{\rm dim} \ 512 \\ {\rm head} \ 8\end{array} \!\!\! \right ] \!\!\times\! 3$ & None & $\left[\!\!\! \begin{array}{c} {\rm \ win}  \ 7\!\times\! 7\\{\rm dim} \ 512 \\ {\rm head} \ 8\end{array} \!\!\! \right ] \!\!\times\! 2$ & None & $\left[\!\!\! \begin{array}{c} {\rm \ win}  \ 7\!\times\! 7\\{\rm dim} \ 512 \\ {\rm head} \ 8\end{array} \!\!\! \right ] \!\!\times\! 3$\\
    \thickhline
    \end{tabular}}
    \caption{Architectures of Slide-PVTv2 models (Part2).}
    \label{model_PVTv2-2}
\end{table*}

\begin{table*}\small
    \centering
    \setlength{\tabcolsep}{1mm}{
    \renewcommand\arraystretch{1.15}
    \begin{tabular}{c|c|c|c|c|c|c|c}
    \thickhline
    \multirow{2}*{stage} & \multirow{2}*{output} & \multicolumn{2}{c|}{Slide-Swin-T} & \multicolumn{2}{c|}{Slide-Swin-S} & \multicolumn{2}{c}{Slide-Swin-B}\\
    \cline{3-8}
    & & \textbf{Slide Attention} & Swin Block & \textbf{Slide Attention} & Swin Block& \textbf{Slide Attention} & Swin Block\\
    \hline
    \multirow{4}*{res1} & \multirow{4}*{$56\times 56$} & \multicolumn{2}{c|}{concat $4\times 4$, 96, LN} & \multicolumn{2}{c|}{concat $4\times 4$, 96, LN} & \multicolumn{2}{c}{concat $4\times 4$, 128, LN}\\
    \cline{3-8}
    && $\left[\!\!\! \begin{array}{c} {\rm \ win}  \ 3\!\times\! 3\\{\rm dim} \ 96 \\ {\rm head} \ 3\end{array} \!\!\! \right ] \!\!\times\! 2$ & None & $\left[\!\!\! \begin{array}{c} {\rm \ win}  \ 3\!\times\! 3\\{\rm dim} \ 96 \\ {\rm head} \ 3\end{array} \!\!\! \right ] \!\!\times\! 2$ & None & $\left[\!\!\! \begin{array}{c} {\rm \ win}  \ 3\!\times\! 3\\{\rm dim} \ 128 \\ {\rm head} \ 3\end{array} \!\!\! \right ] \!\!\times\! 2$ & None\\
    \hline
    \multirow{4}*{res2} & \multirow{4}*{$28\times 28$} & \multicolumn{2}{c|}{concat $4\times 4$, 192, LN} & \multicolumn{2}{c|}{concat $4\times 4$, 192, LN} & \multicolumn{2}{c}{concat $4\times 4$, 256, LN}\\
    \cline{3-8}
    && $\left[\!\!\! \begin{array}{c} {\rm \ win}  \ 3\!\times\! 3\\{\rm dim} \ 192 \\ {\rm head} \ 6\end{array} \!\!\! \right ] \!\!\times\! 2$ & None & $\left[\!\!\! \begin{array}{c} {\rm \ win}  \ 3\!\times\! 3\\{\rm dim} \ 192 \\ {\rm head} \ 6\end{array} \!\!\! \right ] \!\!\times\! 2$ & None & $\left[\!\!\! \begin{array}{c} {\rm \ win}  \ 3\!\times\! 3\\{\rm dim} \ 256 \\ {\rm head} \ 6\end{array} \!\!\! \right ] \!\!\times\! 2$ & None\\
    \hline
    \multirow{4}*{res3} & \multirow{4}*{$14\times 14$} & \multicolumn{2}{c|}{concat $4\times 4$, 384, LN} & \multicolumn{2}{c|}{concat $4\times 4$, 384, LN} & \multicolumn{2}{c}{concat $4\times 4$, 512, LN}\\
    \cline{3-8}
    && None & $\left[\!\!\! \begin{array}{c} {\rm \ win}  \ 7\!\times\! 7\\{\rm dim} \ 384 \\ {\rm head} \ 12\end{array} \!\!\! \right ] \!\!\times\! 6$ & None & $\left[\!\!\! \begin{array}{c} {\rm \ win}  \ 7\!\times\! 7\\{\rm dim} \ 384 \\ {\rm head} \ 12\end{array} \!\!\! \right ] \!\!\times\! 18$ & None & $\left[\!\!\! \begin{array}{c} {\rm \ win}  \ 7\!\times\! 7\\{\rm dim} \ 512 \\ {\rm head} \ 12\end{array} \!\!\! \right ] \!\!\times\! 18$\\
    \hline
    \multirow{4}*{res4} & \multirow{4}*{$7\times 7$} & \multicolumn{2}{c|}{concat $4\times 4$, 768, LN} & \multicolumn{2}{c|}{concat $4\times 4$, 768, LN} & \multicolumn{2}{c}{concat $4\times 4$, 1024, LN}\\
    \cline{3-8}
    & & None& $\left[\!\!\! \begin{array}{c} {\rm \ win}  \ 7\!\times\! 7\\{\rm dim} \ 768 \\ {\rm head} \ 24\end{array} \!\!\! \right ] \!\!\times\! 2$ & None & $\left[\!\!\! \begin{array}{c} {\rm \ win}  \ 7\!\times\! 7\\{\rm dim} \ 768 \\ {\rm head} \ 24\end{array} \!\!\! \right ] \!\!\times\! 2$ & None & $\left[\!\!\! \begin{array}{c} {\rm \ win}  \ 7\!\times\! 7\\{\rm dim} \ 1024 \\ {\rm head} \ 24\end{array} \!\!\! \right ] \!\!\times\! 2$\\
    \thickhline
    \end{tabular}}
    \caption{Architectures of Slide-Swin models.}
    \label{model_swin}
\end{table*}

\begin{table*}\small
    \centering
    \setlength{\tabcolsep}{1mm}{
    \renewcommand\arraystretch{1.15}
    \begin{tabular}{c|c|c|c|c|c|c|c}
    \thickhline
    \multirow{2}*{stage} & \multirow{2}*{output} & \multicolumn{2}{c|}{Slide-CSwin-T} & \multicolumn{2}{c|}{Slide-CSwin-S} & \multicolumn{2}{c}{Slide-CSwin-B}\\
    \cline{3-8}
    & & \textbf{Slide Attention} & CSwin Block & \textbf{Slide Attention} & CSwin Block& \textbf{Slide Attention} & CSwin Block\\
    \hline
    \multirow{4}*{res1} & \multirow{4}*{$56\times 56$} & \multicolumn{4}{c|}{Conv7×7, stride=4, 64, LN} & \multicolumn{2}{c}{Conv7×7, stride=4, 96, LN} \\
    \cline{3-8}
    && $\left[\!\!\! \begin{array}{c} {\rm \ win}  \ 3\!\times\! 3\\{\rm dim} \ 64 \\ {\rm head} \ 2\end{array} \!\!\! \right ] \!\!\times\! 1$ & None & $\left[\!\!\! \begin{array}{c} {\rm \ win}  \ 3\!\times\! 3\\{\rm dim} \ 64 \\ {\rm head} \ 2\end{array} \!\!\! \right ] \!\!\times\! 2$ & None & $\left[\!\!\! \begin{array}{c} {\rm \ win}  \ 3\!\times\! 3\\{\rm dim} \ 96 \\ {\rm head} \ 4\end{array} \!\!\! \right ] \!\!\times\! 2$ & None\\
    \hline
    \multirow{4}*{res2} & \multirow{4}*{$28\times 28$} & \multicolumn{4}{c|}{Conv7×7, stride=4, 128, LN} & \multicolumn{2}{c}{Conv7×7, stride=4, 192, LN} \\
    \cline{3-8}
    && $\left[\!\!\! \begin{array}{c} {\rm \ win}  \ 3\!\times\! 3\\{\rm dim} \ 128 \\ {\rm head} \ 4\end{array} \!\!\! \right ] \!\!\times\! 2$ & None & $\left[\!\!\! \begin{array}{c} {\rm \ win}  \ 3\!\times\! 3\\{\rm dim} \ 128 \\ {\rm head} \ 4\end{array} \!\!\! \right ] \!\!\times\! 4$ & None & $\left[\!\!\! \begin{array}{c} {\rm \ win}  \ 3\!\times\! 3\\{\rm dim} \ 192 \\ {\rm head} \ 8\end{array} \!\!\! \right ] \!\!\times\! 4$ & None\\
    \hline
    \multirow{4}*{res3} & \multirow{4}*{$14\times 14$} & \multicolumn{4}{c|}{Conv7×7, stride=4, 256, LN} & \multicolumn{2}{c}{Conv7×7, stride=384, LN} \\
    \cline{3-8}
    && $\left[\!\!\! \begin{array}{c} {\rm \ win}  \ 3\!\times\! 3\\{\rm dim} \ 256 \\ {\rm head} \ 8\end{array} \!\!\! \right ] \!\!\times\! 5$ & $\left[\!\!\! \begin{array}{c} {\rm \ win}  \ 7\!\times\! 7\\{\rm dim} \ 256 \\ {\rm head} \ 8\end{array} \!\!\! \right ] \!\!\times\! 16$ &  $\left[\!\!\! \begin{array}{c} {\rm \ win}  \ 3\!\times\! 3\\{\rm dim} \ 256 \\ {\rm head} \ 8\end{array} \!\!\! \right ] \!\!\times\! 8$ & $\left[\!\!\! \begin{array}{c} {\rm \ win}  \ 7\!\times\! 7\\{\rm dim} \ 256 \\ {\rm head} \ 8\end{array} \!\!\! \right ] \!\!\times\! 24$ & $\left[\!\!\! \begin{array}{c} {\rm \ win}  \ 3\!\times\! 3\\{\rm dim} \ 384 \\ {\rm head} \ 16\end{array} \!\!\! \right ] \!\!\times\! 14$ & $\left[\!\!\! \begin{array}{c} {\rm \ win}  \ 7\!\times\! 7\\{\rm dim} \ 384 \\ {\rm head} \ 16\end{array} \!\!\! \right ] \!\!\times\! 18$\\
    \hline
    \multirow{4}*{res4} & \multirow{4}*{$7\times 7$} & \multicolumn{4}{c|}{Conv7×7, stride=4, 512, LN} & \multicolumn{2}{c}{Conv7×7, stride=4, 768, LN} \\
    \cline{3-8}
    & & None& $\left[\!\!\! \begin{array}{c} {\rm \ win}  \ 7\!\times\! 7\\{\rm dim} \ 512 \\ {\rm head} \ 16\end{array} \!\!\! \right ] \!\!\times\! 1$ & None & $\left[\!\!\! \begin{array}{c} {\rm \ win}  \ 7\!\times\! 7\\{\rm dim} \ 512 \\ {\rm head} \ 16\end{array} \!\!\! \right ] \!\!\times\! 2$ & None & $\left[\!\!\! \begin{array}{c} {\rm \ win}  \ 7\!\times\! 7\\{\rm dim} \ 768 \\ {\rm head} \ 32\end{array} \!\!\! \right ] \!\!\times\! 2$\\
    \thickhline
    \end{tabular}}
    \caption{Architectures of Slide-CSwin models.}
    \label{model_cswin}
\end{table*}

\begin{table*}\small
    \centering
    \setlength{\tabcolsep}{1mm}{
    \renewcommand\arraystretch{1.15}
    \begin{tabular}{c|c|c|c|c|c|c|c}
    \thickhline
    \multirow{2}*{stage} & \multirow{2}*{output} & \multicolumn{2}{c|}{Slide-NAT-Mini} & \multicolumn{2}{c|}{Slide-NAT-Tiny} & \multicolumn{2}{c}{Slide-NAT-Small}\\
    \cline{3-8}
    & & \textbf{Slide Attention} & Swin Block & \textbf{Slide Attention} & Swin Block& \textbf{Slide Attention} & Swin Block\\
    \hline
    \multirow{4}*{res1} & \multirow{4}*{$56\times 56$} & \multicolumn{4}{c|}{2 * Conv3×3, stride=2, 64, LN} & \multicolumn{2}{c}{2 * Conv3×3, stride=2, 96, LN} \\
    \cline{3-8}
    && $\left[\!\!\! \begin{array}{c} {\rm \ win}  \ 3\!\times\! 3\\{\rm dim} \ 64 \\ {\rm head} \ 2\end{array} \!\!\! \right ] \!\!\times\! 3$ & None & $\left[\!\!\! \begin{array}{c} {\rm \ win}  \ 3\!\times\! 3\\{\rm dim} \ 64 \\ {\rm head} \ 2\end{array} \!\!\! \right ] \!\!\times\! 3$ & None & $\left[\!\!\! \begin{array}{c} {\rm \ win}  \ 3\!\times\! 3\\{\rm dim} \ 96 \\ {\rm head} \ 3\end{array} \!\!\! \right ] \!\!\times\! 3$ & None\\
    \hline
    \multirow{4}*{res2} & \multirow{4}*{$28\times 28$} & \multicolumn{4}{c|}{Conv3×3, stride=2, 128, LN} & \multicolumn{2}{c}{Conv3×3, stride=2, 192, LN} \\
    \cline{3-8}
    && $\left[\!\!\! \begin{array}{c} {\rm \ win}  \ 3\!\times\! 3\\{\rm dim} \ 128 \\ {\rm head} \ 4\end{array} \!\!\! \right ] \!\!\times\! 4$ & None & $\left[\!\!\! \begin{array}{c} {\rm \ win}  \ 3\!\times\! 3\\{\rm dim} \ 128 \\ {\rm head} \ 4\end{array} \!\!\! \right ] \!\!\times\! 4$ & None & $\left[\!\!\! \begin{array}{c} {\rm \ win}  \ 3\!\times\! 3\\{\rm dim} \ 192 \\ {\rm head} \ 6\end{array} \!\!\! \right ] \!\!\times\! 4$ & None\\
    \hline
    \multirow{4}*{res3} & \multirow{4}*{$14\times 14$} & \multicolumn{4}{c|}{Conv3×3, stride=2, 256, LN} & \multicolumn{2}{c}{Conv3×3, stride=2, 384, LN} \\
    \cline{3-8}
    && None & $\left[\!\!\! \begin{array}{c} {\rm \ win}  \ 7\!\times\! 7\\{\rm dim} \ 256 \\ {\rm head} \ 8\end{array} \!\!\! \right ] \!\!\times\! 6$ &  $\left[\!\!\! \begin{array}{c} {\rm \ win}  \ 3\!\times\! 3\\{\rm dim} \ 256 \\ {\rm head} \ 8\end{array} \!\!\! \right ] \!\!\times\! 10$ & $\left[\!\!\! \begin{array}{c} {\rm \ win}  \ 7\!\times\! 7\\{\rm dim} \ 256 \\ {\rm head} \ 8\end{array} \!\!\! \right ] \!\!\times\! 8$ & $\left[\!\!\! \begin{array}{c} {\rm \ win}  \ 3\!\times\! 3\\{\rm dim} \ 384 \\ {\rm head} \ 12\end{array} \!\!\! \right ] \!\!\times\! 10$ & $\left[\!\!\! \begin{array}{c} {\rm \ win}  \ 7\!\times\! 7\\{\rm dim} \ 384 \\ {\rm head} \ 12\end{array} \!\!\! \right ] \!\!\times\! 8$\\
    \hline
    \multirow{4}*{res4} & \multirow{4}*{$7\times 7$} & \multicolumn{4}{c|}{Conv3×3, stride=2, 512, LN} & \multicolumn{2}{c}{Conv3×3, stride=2, 768, LN} \\
    \cline{3-8}
    & & None& $\left[\!\!\! \begin{array}{c} {\rm \ win}  \ 7\!\times\! 7\\{\rm dim} \ 512 \\ {\rm head} \ 16\end{array} \!\!\! \right ] \!\!\times\! 5$ & None & $\left[\!\!\! \begin{array}{c} {\rm \ win}  \ 7\!\times\! 7\\{\rm dim} \ 512 \\ {\rm head} \ 16\end{array} \!\!\! \right ] \!\!\times\! 5$ & None & $\left[\!\!\! \begin{array}{c} {\rm \ win}  \ 7\!\times\! 7\\{\rm dim} \ 768 \\ {\rm head} \ 24\end{array} \!\!\! \right ] \!\!\times\! 5$\\
    \thickhline
    \end{tabular}}
    \caption{Architectures of Slide-NAT models.}
    \label{model_nat}
\end{table*}

{\small
\bibliographystyle{ieee_fullname}
\bibliography{egbib}
}

\end{document}